\documentclass[sn-mathphys,Numbered]{sn-jnl}

\usepackage{graphicx}%
\usepackage{multirow}%
\usepackage{amsmath,amssymb,amsfonts}%
\usepackage{amsthm}%
\usepackage{mathrsfs}%
\usepackage[title]{appendix}%
\usepackage{xcolor}%
\usepackage{textcomp}%
\usepackage{manyfoot}%
\usepackage{booktabs}%
\usepackage{algorithmicx}%
\usepackage[noend]{algpseudocode}%
\usepackage{listings}%
\usepackage{graphicx}%
\usepackage{diagbox}
\usepackage{pifont}
\usepackage[ruled,vlined,linesnumbered,boxed,commentsnumbered]{algorithm2e}

\theoremstyle{thmstyleone}%
\theoremstyle{thmstyletwo}%

\theoremstyle{thmstylethree}%

\begin{document}

\title[Article Title]{Physics-Driven Spectrum-Consistent Federated Learning for Palmprint Verification}



\author[1]{\fnm{Ziyuan} \sur{Yang}}\email{cziyuanyang@gmail.com}
\author[2]{\fnm{Andrew Beng Jin} \sur{Teoh}}\email{bjteoh@yonsei.ac.kr}
\author[3]{\fnm{Bob} \sur{Zhang}}\email{bobzhang@um.edu.mo}
\author[4]{\fnm{Lu} \sur{Leng}}\email{leng@nchu.edu.cn}
\author*[5]{\fnm{Yi} \sur{Zhang}}\email{yzhang@scu.edu.cn}

\affil[1]{\orgdiv{College of Computer Science}, \orgname{Sichuan University}, \orgaddress{\city{Chengdu}, \country{China}}}

\affil[2]{\orgdiv{School of Electrical and Electronic Engineering, College of Engineering}, \orgname{Yonsei University}, \orgaddress{\city{Seoul}, \country{Republic of Korea}}}

\affil[3]{\orgdiv{Pattern Analysis and Machine Intelligence Group, Department of Computer and Information Science}, \orgname{University of Macau}, \orgaddress{\city{Macau}, \country{China}}}

\affil[4]{\orgdiv{School of Software}, \orgname{Nanchang Hangkong University}, \orgaddress{\city{Nanchang}, \country{China}}}

\affil*[5]{\orgdiv{School of Cyber Science and Engineering}, \orgname{Sichuan University}, \orgaddress{\city{Chengdu}, \country{China}}}


\abstract{
Palmprint as biometrics has gained increasing attention recently due to its discriminative ability and robustness. However, existing methods mainly improve palmprint verification within one spectrum, which is challenging to verify across different spectrums. Additionally, in distributed server-client-based deployment, palmprint verification systems predominantly necessitate clients to transmit private data for model training on the centralized server, thereby engendering privacy apprehensions. To alleviate the above issues, in this paper, we propose a \textbf{p}hysics-driven \textbf{s}pectrum-consistent \textbf{fed}erated learning method for \textbf{palm}print verification, dubbed as PSFed-Palm. PSFed-Palm draws upon the inherent physical properties of distinct wavelength spectrums, wherein images acquired under similar wavelengths display heightened resemblances. Our approach first partitions clients into short- and long-spectrum groups according to the wavelength range of their local spectrum images. Subsequently, we introduce anchor models for short- and long-spectrum, which constrain the optimization directions of local models associated with long- and short-spectrum images. Specifically, a spectrum-consistent loss that enforces the model parameters and feature representation to align with their corresponding anchor models is designed. Finally, we impose constraints on the local models to ensure their consistency with the global model, effectively preventing model drift. This measure guarantees spectrum consistency while protecting data privacy, as there is no need to share local data. Extensive experiments are conducted to validate the efficacy of our proposed PSFed-Palm approach. The proposed PSFed-Palm demonstrates compelling performance despite only a limited number of training data. The codes will be released at https://github.com/Zi-YuanYang/PSFed-Palm.
}
\keywords{Biometrics, palmprint verification, spectrum-consistent federated learning, privacy-preserving}



\maketitle

\section{Introduction}\label{sec1}

Biometrics technology plays a crucial role in modern identity management and security, offering dependable methods for authentication and identification based on distinctive biological features. While face and iris biometrics have been widely employed, they have exposed several limitations in real-world applications \cite{han2022personalized}. For example, wearing a mask or a goggle during an epidemic would unavoidably reduce recognition performances.
Hence, increasing efforts have recently been devoted to developing palmprint-based biometrics, which has emerged as a promising research field~\cite{gomez2021biometrics}. 

Over the past few decades, significant progress has been made in developing accurate and reliable methods for palmprint biometrics. For instance, Zhang \textit{et al.}~\cite{zhang2003online} proposed PalmCode, which applied Gabor filters to extract texture features to verify and achieve satisfactory performance. Inspired by this work, many variants have been proposed in this field~\cite{xu2018drcc}. However, these methods heavily rely on prior knowledge and expertise, potentially limiting their verification performance and robustness.

Integrating deep learning (DL) into palmprint biometrics has recently garnered significant attention from researchers seeking to overcome its inherent limitations. Particularly noteworthy is the emerging trend of fusing DL with Gabor filters, facilitating the development of trainable Gabor filters capable of autonomously extracting discriminative texture features. These novel approaches have yielded promising performances in palmprint biometrics~\cite{liang2021compnet,yang2023co3net}.

Despite the satisfactory performance of numerous methods, a common assumption is that palmprint images are acquired under homogeneous environmental conditions within a single spectrum. However, there is significant heterogeneity among images acquired from different spectrums. For instance, unlike visible light, near-infrared (NIR) imaging can capture palm vein information through the palm surface, making it valuable for nighttime palmprint image acquisition~\cite{zhang2009online}. To address this challenge, a potential solution is to achieve consistency across different spectrum palmprint templates~\cite{dong2022co}. This would enable companies to use suitable devices for acquiring users' palmprint images based on different application scenarios without repeated registration. 


In the context of practical distributed client-server palmprint biometrics deployment, end-users can enroll themselves with diverse clients, encompassing different bank branches, retail stores, and other relevant entities. Most existing systems require clients to upload their data to a centralized server for training~\cite{fei2020feature}. However, this oversight neglects the sensitive and private information present in palmprint images of users, giving rise to legitimate privacy apprehensions. Besides, the data in the channel can be hacked so that private data can be utilized to threaten the verification systems. Hence, it is challenging to implement these methods in practice due to the multiple strict privacy regulations and potential security concerns. Collaboratively training a powerful and robust model among clients, capable of minimizing the feature disparity inherent in heterogeneous spectrum palmprint images without data leakage, poses an intriguing and challenging problem.

To address the above challenges, we propose a physics-driven spectrum-consistent federated learning framework for palmprint verification called PSFed-Palm. This framework is inspired by the physical characteristics of various wavelength spectrums. The heterogeneity arising from diverse acquisition spectrums can induce disparate optimization directions for local models during the training process, thereby giving rise to considerable discrepancies between templates derived from different spectrums. To mitigate this issue, we first divide existing spectrums into short- and long-spectrum groups based on their corresponding wavelengths. We assume that models belonging to the same group would extract similar templates. This assumption is based on the fundamental physical phenomenon that images acquired using similar wavelengths tend to demonstrate heightened resemblance. Accordingly, we put forward anchor models for short- and long-spectrum groups by combining client results within each group. To ensure spectrum consistency across groups without data sharing, we devise a spectrum-consistent loss that prevents local models in one group from deviating from the parameters and representations of the anchor model from another group. 

We must mention that significant differences exist between our and previous works, such as~\cite{dong2022co}. In~\cite{dong2022co}, different spectral images of the same palm were fed into the network, and specific loss functions were designed to achieve spectrum consistency. In contrast, our method takes a model-based approach. It leverages the knowledge embedded within the anchor model, which is trained on different spectrum data than the local spectrum. This knowledge is utilized to achieve spectrum consistency in our approach, eliminating the need for clients to collect diverse spectral images of the same palm simultaneously.

In addition, this work aims to generate a distinct model for each client in a server-client architecture. To achieve this, the server aggregates all individual local models to construct the global model. In each communication round, the server aggregates local models, resulting in three distinct models: two anchor models and one global model. To sum up, the incorporation of anchor models, along with spectrum-consistent loss, serves to prevent local models from deviating significantly from the anchor model of another designated group; hence, the proposed PSFed-Palm can ensure the templates generated from different spectrums become consistent, and the training process is stable. Moreover, leveraging the distributed training paradigm, the global and anchor models acquire comprehensive knowledge from all clients and spectrum groups. PSFed-Palm attains commendable performance even with limited local single-spectrum data. This capability originates from utilizing the acquired knowledge effectively, mitigating overfitting issues commonly encountered by local models during training with limited data or local spectra.

  
The main contributions of this paper can be summarized as follows:
\begin{itemize}
    \item We propose a novel physics-driven spectrum-consistent federated learning framework for palmprint verification that can preserve privacy and simultaneously achieve spectrum consistency. To the best of our knowledge, this is the first attempt to establish spectrum consistency for palmprint verification within a distributed learning environment.
    
    \item Inspired by the physical characteristics, we introduce a spectrum-consistent loss to enforce adherence of local models to the models trained on disparate spectra without access to external spectrum data in a single client.

    \item The proposed method preserves privacy by circumventing palmprint data transfer during training. The global and anchor models are exchanged, inherently containing substantially less privacy-sensitive information than the raw data.

    \item We validated the proposed framework's efficacy compared to other federated learning (FL) methods by conducting extensive experiments on publicly available datasets.
    
\end{itemize}


The paper is organized as follows: Section 2 reviews related works, Section 3 elaborates on the proposed method, Section 4 discusses experiments to validate the proposed method, and Section 5 concludes the work and suggests future works.

\section{Related Works}
\subsection{Palmprint Verification}
Palmprint biometrics has gained widespread popularity in various applications due to its user-friendliness, privacy, and high discriminability. Palmprint verification can be broadly categorized into four groups: subspace-based, statistic-based, coding-based, and DL-based methods~\cite{zhong2019decade}. Subspace-based methods attempt to design projection formulations to map palmprint images onto a lower-dimensional subspace. Statistic-based methods typically involve steps, extracting features from palmprint images and employing statistical techniques to extract discriminative information~\cite{zhang2018combining}.

Coding-based methods aim to extract discriminative texture features for verifying individuals~\cite{fei2020feature}. For instance, Guo \textit{et al.}~\cite{guo2009palmprint} extended PalmCode and proposed Binary Orientation Co-occurrence Vector (BOCV), which utilized six Gabor filters along different directions to extract magnitude features and then encoded them for matching. Inspired by this work, Yang \textit{et al.}~\cite{yang20212TCC} proposed to extract 2nd-order features and combine them with 1st-order features to achieve better performance. Accordingly, several magnitude feature-based methods have been proposed following this vein~\cite{zhang2012fragile, sun2005ordinal, yang2021extreme, kong2006palmprint}.
However, magnitude features are sensitive to illumination changes, which impede the robustness of these methods~\cite{fei2018feature}. Kong \textit{et al.}~\cite{kong2003palmprint} first introduced the competition mechanism into palmprint biometrics to propose Competitive Code (Comp Code) and achieved robust performance. Motivated by the success of Comp Code, numerous competition mechanism-based methods have been proposed in the past decades \cite{jia2008palmprint}, such as Half-Orientation Code (HOC)~\cite{fei2016half}, Discriminative Robust Competitive Code (DRCC)~\cite{xu2018drcc}, and Double Orientation Code (DOC)~\cite{fei2016double}.

However, the above methods highly rely on prior knowledge. Inspired by the significant achievements of deep learning (DL) in computer vision~\cite{jiang2023adversarial}, researchers have exhibited great interest in integrating DL techniques into palmprint biometrics~\cite{zhao2022structure}. For example, Matkowski \textit{et al.}~\cite{matkowski2019palmprint} proposed a DL-based palmprint recognition method in an uncontrolled and uncooperative environment. Chai \textit{et al.}~\cite{chai2019boosting} presented a convolutional neural network (CNN) to achieve improved palmprint recognition performance by integrating soft biometric information. Genovese \textit{et al.}~\cite{genovese2019palmnet} proposed an unsupervised learning method named PalmNet. Furthermore, Zhang \textit{et al.}~\cite{zhong2018palm} introduced a deep hashing palm network (DHPN), which extended from deep hashing networks~\cite{zhu2016deep} to palmprint biometrics. Building upon DHPN, Wu \textit{et al.}~\cite{wu2021palmprint} incorporated spatial transformer modules, leading to a notable performance improvement. In a recent endeavor named CompNet~\cite{liang2021compnet}, the extension of Gabor filters into a learnable form was explored to reduce dependence on prior knowledge. Motivated by the insights from this work, an approach named CO3Net~\cite{yang2023co3net} was introduced, focusing on utilizing coordinate information and feature space optimization to attain improved performance.

Existing methods have achieved reasonable performance, but only under similar environmental conditions. There have been some recent proposals to address this problem. For example, Cho \textit{et al.}~\cite{cho2019extraction} proposed a method for converting visible images to invisible spectrum palmprint images. They extended this approach to develop a cross-matching technique that utilizes palm vein and palmprint images for identity verification~\cite{cho2021palm}. Additionally, \cite{dong2022co} and \cite{su2023learning} endeavored to design consistent loss functions to extract templates from diverse spectrum palmprint images. Nevertheless, the above methods assume that a client can access all spectrum palmprint images from various sources simultaneously, and some approaches only consider the consistency between specific spectrums. Consequently, these methods face challenges in practical implementation due to their oversight of privacy concerns.

\subsection{Federated Learning}

FL is one of the most popular distributed cooperative machine learning paradigms, which can train a global-shared network at each client in parallel. Then the local gradients would be transferred to the server for aggregation~\cite{wicaksana2022customized}. FedAvg~\cite{mcmahan2017communication} is a typical method, which weighted aggregates the gradients of local models after each communication round in the server. Inspired by this work, Li \textit{et al.}~\cite{li2020federated} proposed FedProx, which designed a proximal regularization to achieve a stable training process based on FedAvg. Besides, Li \textit{et al.}~\cite{li2021fedbn} personalized batch normalization layers in local models without aggregating them to achieve personalization characteristics. Arivazhagan \textit{et al.}~\cite{arivazhagan2019federated} proposed aggregating most layers and personalizing specific layers to address the heterogeneity problem. In \cite{wu2022communication}, each client has two models, including a small and a large model, and only the small model is uploaded to the server for aggregation to achieve communication efficiency. In local training, the large model utilizes the small model to learn robust knowledge.
Similarly, Liang \textit{et al.}~\cite{liang2020think} proposed simultaneously learning personalized and global models. Inspired by the physical imaging process of CT scanning, Yang \textit{et al.}~\cite{yang2022hypernetwork} proposed a physics-driven FL for CT imaging. Shao and Zhong~\cite{shao2020towards} introduced FL into palmprint recognition, but they assume that clients share a public dataset in the server, which could compromise privacy.

\section{Physics-Driven Spectrum-Consistent Federated Learning}
\subsection{Overview}

\begin{figure}[]
    \centerline{\includegraphics[width=\textwidth]{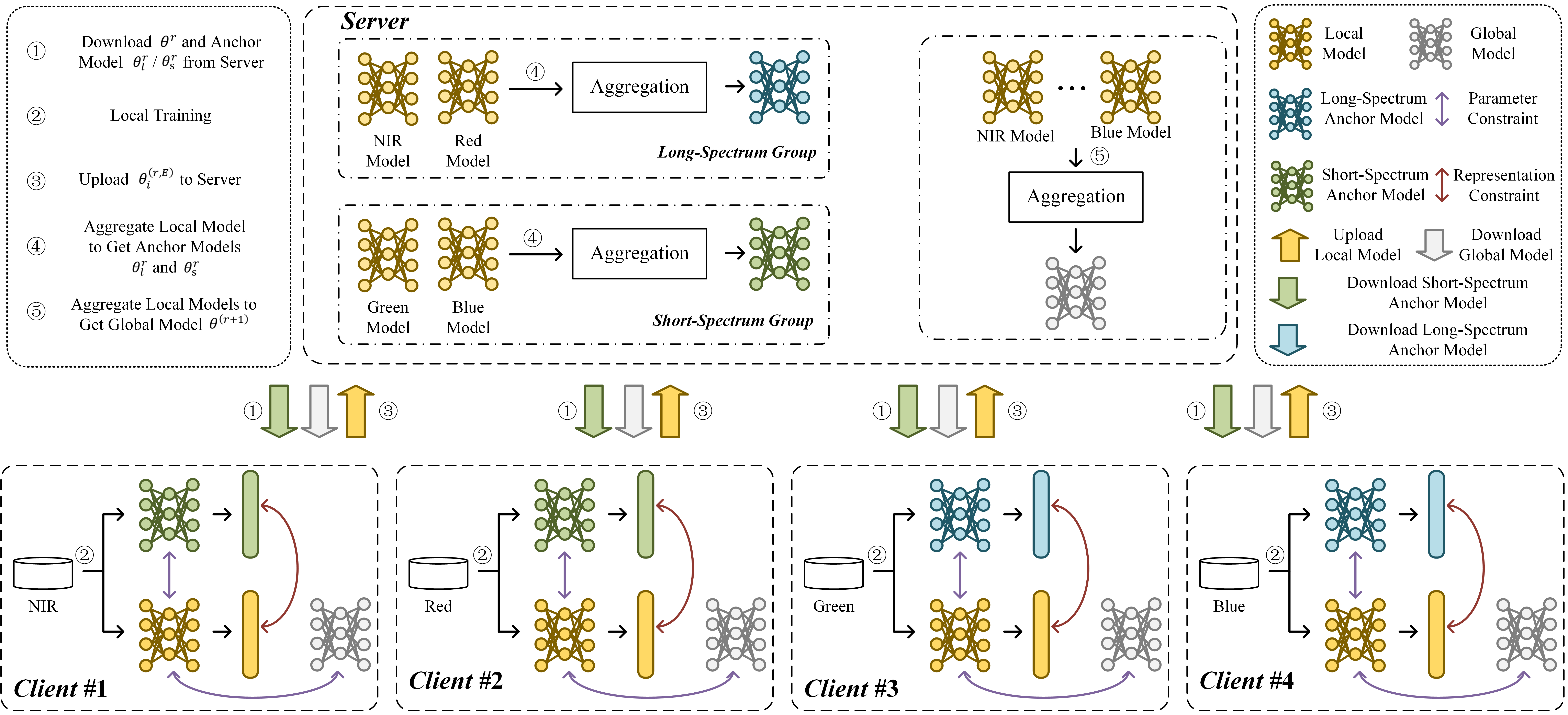}}
    \caption{The overview of the proposed PSFed-Palm.}
    \label{fig:fraemwork}
\end{figure}

Current palmprint verification methods neglect privacy concerns and prioritize performance in a single spectrum. To alleviate the above two issues, we propose a novel physics-driven spectrum-consistent federated learning (PSFed-Palm) framework, as depicted in Fig.~\ref{fig:fraemwork}. Users' data privacy can be effectively preserved since the data is only stored locally and not shared with other clients or servers. To bridge the gap between different spectrums, we utilize anchor models to prevent local models from deviating from the other group's anchor model. Additionally, we use the constraint between local and global models to avoid shifts in the model.

\begin{figure}[!t]
\vspace{8pt}
	\centering
	\begin{minipage}[t]{0.24\textwidth}
	\includegraphics[width=\textwidth]{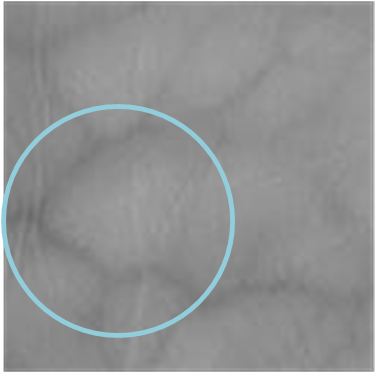}
	\centerline{(a)}
	\end{minipage}
	\begin{minipage}[t]{0.24\textwidth}
	\includegraphics[width=\textwidth]{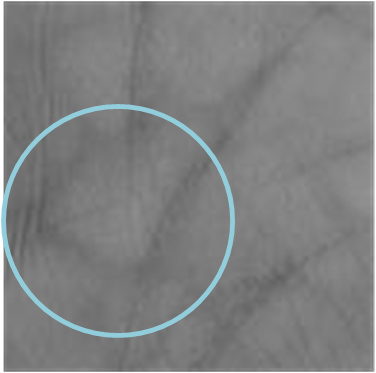}
	\centerline{(b)}
	\end{minipage}
	\begin{minipage}[t]{0.24\textwidth}
	\centering
	\includegraphics[width=\textwidth]{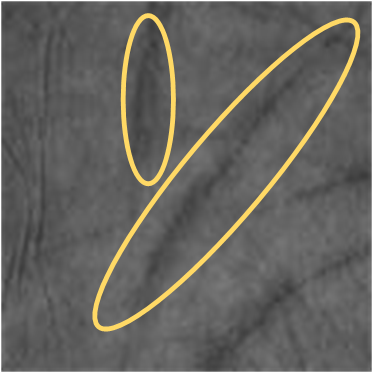}
	\centerline{(c)}
	\end{minipage}
	\begin{minipage}[t]{0.24\textwidth}
	\includegraphics[width=\textwidth]{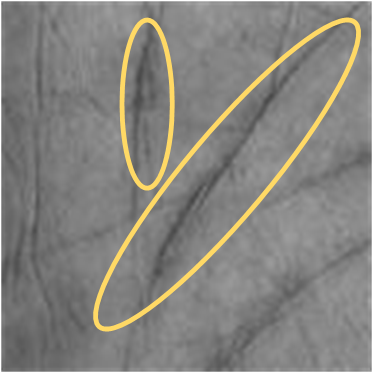}
	\centerline{(d)}
	\end{minipage}
 \vspace{5pt}
\caption{Acquired palmprint samples through different spectra. (a)-(d) represent NIR, Red, Green, and Blue, respectively. The yellow circles indicate textures, while the blue circles indicate veins.}
\label{fig: sample}
\end{figure}

\begin{table}[]
\caption{The wavelength ranges of different spectrums (nm).}
\begin{tabular}{@{}ccccc@{}}
\toprule
  & NIR  & Red & Green & Blue\\
\midrule
Wavelength Range    & 760–900 & 630–690 & 520–600 & 450–520  \\
\botrule
\end{tabular}
\label{tab: wavelength}
\end{table}

\subsection{Motivation}


The acquired palm images display significant heterogeneity due to the varying physical properties of wavelengths in different spectra, as depicted in Fig.~\ref{fig: sample}. This discrepancy can be attributed to the variations in wavelengths across different spectrums~\cite{dejonge2016assessing}, as shown in Tab.~\ref{tab: wavelength}. The varying wavelengths result in unique characteristics such as penetration, absorption, and reflection, leading to distinct images. For instance, vein images can be acquired through NIR light due to higher hemoglobin absorption in blood than in the skin and other tissues. NIR images exhibit distinct characteristics as the wavelength increases for palmprint verification. Specifically, tissue penetration is enhanced, resulting in a weakened representation of texture information, while venous information becomes more pronounced. Conversely, shorter wavelengths have a lower tissue-penetrating ability, more substantial texture information, and less prominent venous information, as shown in Fig.~\ref{fig: sample}. Therefore, different spectral palmprint images exhibit significant heterogeneity, impeding the optimization process of the recognition network.

FL is a promising distributed learning paradigm that effectively preserves privacy. However, the heterogeneity problem between clients remains an open challenge. The variations in the physical characteristics of different spectrums pose significant challenges in palmprint verification. This paper provides a practical scenario assuming that each client only contains one spectrum. While previous methods aim for spectrum consistency~\cite{dong2022co}, they fail to account for the complexities of distributed storage and learning scenarios.

\subsection{Training}
Based on the mentioned phenomena, we divide the clients into two groups based on their wavelength ranges: the short-spectrum group (Green and Blue) and the long-spectrum group (Red and NIR). Notably, the images acquired in the long-spectrum group contain vein information, while those in the short-spectrum group predominantly contain texture information.

In contrast to conventional FL, which directly aggregates local models to obtain a global model after each communication round, and results in a lack of spectrum consistency and significant model drift, our approach aggregates two anchor models for the respective groups alongside the conventional aggregated global model. Specifically, in our method, each client uploads its local model $\mathcal{M}$ to the server after each communication round. $\mathcal{M}$ can be any DL-based palmprint verification model.

For the sake of simplicity, we define $\mathcal{M}_{NIR}$, $\mathcal{M}_{R}$, $\mathcal{M}_{G}$, and $\mathcal{M}_{B}$ as the weighted aggregated models from local models trained by NIR, Red, Green, and Blue spectrum data, respectively, which are parameterized by $\theta_{NIR}$, $\theta_{R}$, $\theta_{G}$, and $\theta_{B}$. This aggregation process takes place on the server.

Based on the different spectrum types, the server categorizes the client models into two distinct groups, denoted as $G_s = \{{\mathcal{M}_{G},\mathcal{M}_{B}}\}$ and $G_l=\{{\mathcal{M}_{R},\mathcal{M}_{NIR}}\}$. $G_s$ and $G_l$ represent the short and long-spectrum models, respectively. Subsequently, the server performs model aggregation within $G_s$ and $G_l$ to derive the short and long-spectrum group anchor models. This process can be formulated as follows:
\begin{equation}
    \theta_s = 1/2 \times \theta_G + 1/2 \times \theta_B
    \label{eq:thetas}
\end{equation}
 and
 \begin{equation}
    \theta_{l} = 1/2\times \theta_{NIR} + 1/2\times \theta_{R}, 
    \label{eq:thetal}
\end{equation}
where $\theta_s$ and $\theta_l$ denote the parameters of $\mathcal{M}_{s}$ and $\mathcal{M}_{l}$, which are the anchor models of short- and long-spectrum group, respectively.

Besides, the proposed PSFed-Palm aims to achieve spectrum consistency by training a global-shared model. Except for $\theta_{s}$ and $\theta_{l}$, the server must aggregate all models to obtain the global model $\theta_{global}$, which can be formulated as follows:
\begin{equation}
    \theta_{global} = 1/2\times \theta_{s} + 1/2\times \theta_{l}, 
    \label{eq:global}
\end{equation}
where $\theta_{global}$ represents the parameters of global model $\mathcal{M}_{global}$.

Once the server finishes the model aggregation, the corresponding models must be downloaded by the clients for local training. Specifically, clients belonging to the short-spectrum group are required to download $\mathcal{M}_{global}$ and $\mathcal{M}_{l}$, while those in the long-spectrum group download $\mathcal{M}_{global}$ and $\mathcal{M}_{s}$. The anchor model is denoted as $\mathcal{M}_{A}$ for each client from the opposite group. For instance, for a client with green spectrum data, $\mathcal{M}_{A}$ corresponds to $\mathcal{M}_{l}$. Notably, $\mathcal{M}_{A}$ assimilates knowledge from models trained on data with significant wavelength differences, thereby differing from the locally trained model.

\begin{figure}
    \centering
    \includegraphics[width=0.9\textwidth]{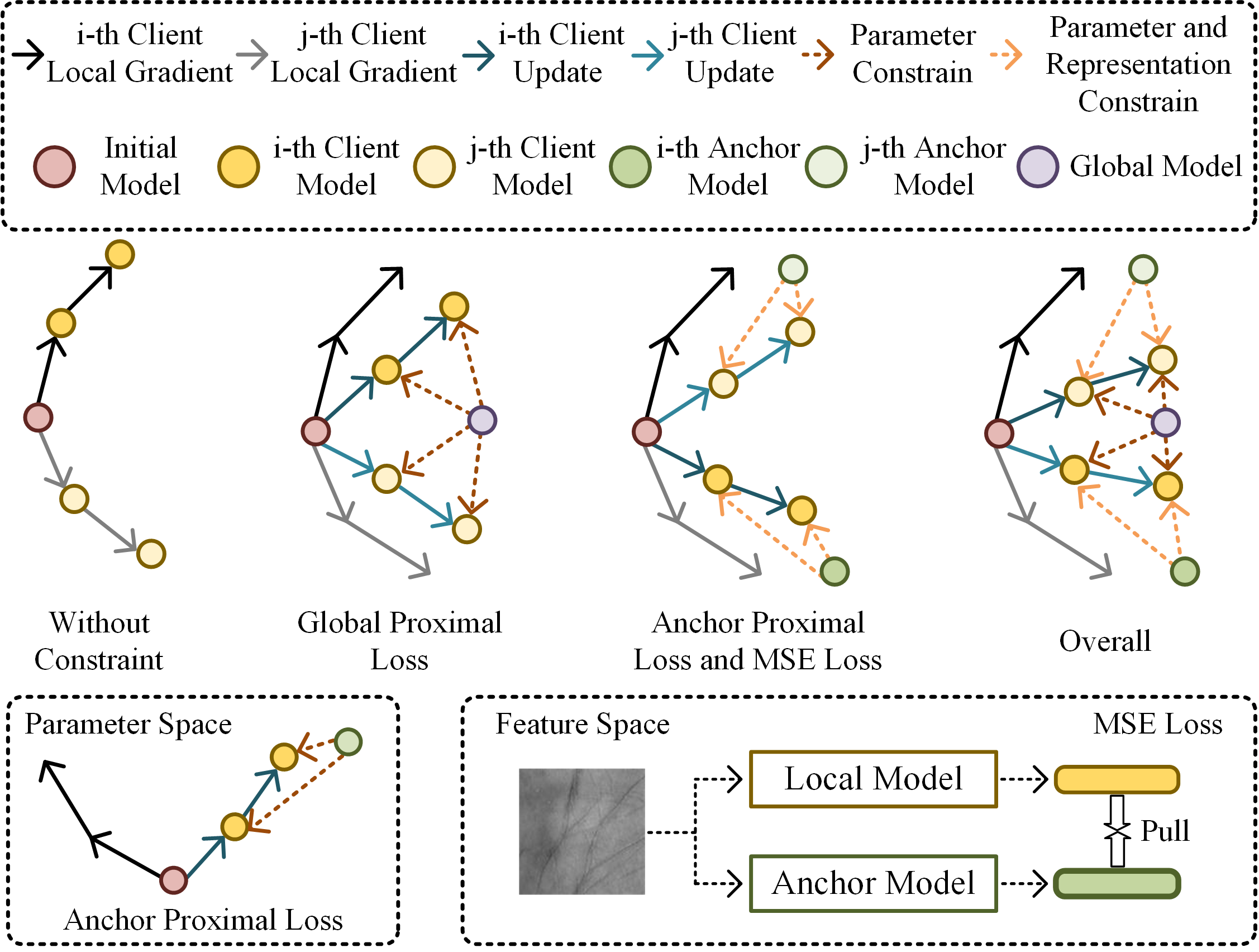}
    \caption{The roles played by different losses in PSFed-Palm framework. "Without Constraint" arises when local training occurs without any constraint losses. In such cases, significant optimization gaps emerge between different local models. "Global Proximal Loss" involves using the global model to impose constraints on the local models, ensuring they do not deviate excessively from the global model. "Anchor Proximal Loss and MSE Loss" is employed to achieve spectrum consistency with the aid of an anchor model. The anchor proximal loss operates in the parameter space, constraining the local model, while the MSE loss works in the representation space, enabling constraint enforcement. "Overall" involves all of the above to ensure a stable training process and facilitates the achievement of spectrum consistency within the PSFed-Palm framework.}
    \label{fig:Loss_illu}
\end{figure}

On the other hand, to mitigate the heterogeneity problem in the absence of constraints during the training process, our framework incorporates two constraint losses: the global proximal loss and the spectrum-consistent loss. The spectrum-consistent loss consists of anchor proximal loss and mean squared error (MSE) loss. The global proximal loss ensures that the local models do not deviate from the global model and stabilizes the training process. However, only the global proximal loss can achieve spectrum consistency. To address this issue, spectrum-consistent loss offers a parameter-level aSpectrum-consistent local model to address this issue by incorporating the anchor model. 

For the parameter constraint, we adopt a classical proximal loss \cite{li2020federated} to constrain the local model and $\mathcal{M}_{A}$, which is formulated as follows:
\begin{equation}
    \mathcal{L}_{prox}(\theta_{local},\theta_{A}) = \frac{\mu}{2}||\theta_{local}-\theta_{A}||_2^2,
    \label{equ:proximal}
\end{equation}
where $\theta_{local}$ and $\theta_{A}$ represent the parameters of $\mathcal{M}_{local}$ and $\mathcal{M}_{A}$, respectively. $\mu$ denotes the temperature parameter.

To obtain a global-shared model, given the inherent data differences across different clients, we utilize Eq.~\eqref{equ:proximal} to impose constraints on the local model to prevent significant deviations from the global model. This inclusion enables a stable training process. To distinguish between these two losses, we refer to them as the global proximal loss, $\mathcal{L}_{prox}^g = \mathcal{L}_{prox}(\theta_i^{(r,e)},\theta_{global})$ and anchor proximal loss $\mathcal{L}_{prox}^A = \mathcal{L}_{prox}(\theta_i^{(r,e)},\theta_A)$, respectively. 

For the representation constraint, we utilize the MSE loss as:
\begin{equation}
    \mathcal{L}_{MSE}=\frac{\tau \sum_{i=1}^{n}(\textbf{v}_A^i-\textbf{v}_l^i)^{2}}{n},
    \label{eq:mse}
\end{equation}
where $\textbf{v}_A \in \mathbb{R}^{1 \times n}$ and $\textbf{v}_l \in \mathbb{R}^{1 \times n}$ denote feature vectors extracted from $\mathcal{M}_{A}$ and $\mathcal{M}_{local}$, respectively.

We illustrate Fig. \ref{fig:Loss_illu} to demonstrate the role of each loss component. $\tau$ denotes the temperature parameter for $\mathcal{L}_{MSE}$.

\begin{algorithm}[t]
\vspace{4pt}
  \caption{Main steps of the PSFed-Palm.}  
  \label{alg:PSFed-Palm}
   \textbf{Definition:} $\theta^t$ denotes the $t$-th communication round of the global model, $\mathcal{M}_i$ denotes the $i$-th client model, $R$ represents the number of communication rounds, $N$ denotes the number of clients, $D^i$ denotes the dataset in the $i$-th client, and $E$ denotes the number of local epochs. $x$ and $y$ denote the palmprint image and corresponding label, respectively.
   
   \textbf{Function Main:} \Comment{Server Executes}
   
    Initialize $\theta^1$. 
    
    \For{round $r=1,2,...,R$}{
    
        \For{client $i=1,2,...,N$ \rm{\textbf{in parallel}}}{
        send $\theta^r$ to $i$-th client
        
        \If{$i$-th client in the short-spectrum group}{
        {$\theta^r_i \gets$ \textbf{Client Local Training}($i, \theta^r, \theta_l$)}  
            }
        \Else
        {$\theta^r_i \gets$ \textbf{Client Local Training}($i, \theta^r, \theta_s$)}
        }        
        $\theta_{NIR}, \theta_{R}, \theta_{G}, \theta_{B} \gets$ Aggregation($\theta^r_1,\theta^r_2,...,\theta^r_N$) \\
                                     \Comment{Based on the spectrum kind of local data}
                                     
        $\theta_s^r \gets$Aggregation($\theta_{G}, \theta_{B}$) \Comment{Based on Eq.~\eqref{eq:thetas}}
        
        $\theta_l^r \gets$Aggregation($\theta_{NIR}, \theta_{R}$) \Comment{Based on Eq.~\eqref{eq:thetal}}
        
        $\theta_{global}^r \gets$ Aggregation($\theta_s, \theta_l$) \Comment{Based on Eq.~\eqref{eq:global}}
        
        $\theta^{r+1} \gets \theta_{global}^r$
    }
    \textbf{Function Client Local Training($i$, $\theta_{global}$, $\theta_A$):} \Comment{Client Executes}

        $\theta_i^{(r,e)}\gets\theta_{global}$
        
        \For{$e=1,2,...,E$}{
        
        \For{$(x, y)$ in $D^i$}{



            $\mathcal{L}_{prox}^A\gets \mathcal{L}_{prox}(\theta_i^{(r,e)},\theta_A)$ \Comment{Based on Eq.~\eqref{equ:proximal}}

            $\mathcal{L}_{prox}^g\gets \mathcal{L}_{prox}(\theta_i^{(r,e)},\theta_{global})$ \Comment{Based on Eq.~\eqref{equ:proximal}}            

            $\mathcal{L}_{MSE}\gets \mathcal{L}_{MSE}(\mathcal{M}_i(x,\theta_i^{(r,e)}),\mathcal{M}_A(x,\theta_A))$ \Comment{Based on Eq.~\eqref{eq:mse}}
            
            Compute the Task Loss $\mathcal{L}_{task}$ based on $(x,y)$ \Comment{Based on Eq.~\eqref{equ:task}}
        
            $\mathcal{L} \gets \mathcal{L}_{task} + \mathcal{L}_{prox}^A + \mathcal{L}_{prox}^{g} + \mathcal{L}_{MSE}$ \Comment{Based on Eq.~\eqref{eq:all}}

            Back-propagation based on $\mathcal{L}$
        }
        }

        \textbf{return} $\theta_i^{(r,E)}$

\end{algorithm}

For palmprint verification, a task loss $\mathcal{L}_{task}$ is required to ensure the network can accurately verify the palmprint. In this paper, we adopt the hybrid loss proposed in~\cite{yang2023co3net} as the task loss for each client, which is the weighted combination of cross-entropy loss and supervised contrastive loss~\cite{khosla2020supervised}, formulated as:

\begin{equation}
    \mathcal{L}_{task} = w_{ce} \times \mathcal{L}_{ce} + w_{con} \times \mathcal{L}_{con},
    \label{equ:task}
\end{equation}

The cross-entropy loss $\mathcal{L}_{ce}$ is given as:
\begin{equation}
    \mathcal{L}_{ce} = -\frac{1}{K}\sum_{i=1}^K\sum_{c=1}^{M} y_{i,c} \log \left(p_{i,c}\right),
\end{equation}
where $K$ and $M$ denote the numbers of samples and classes, respectively. $y_{i,c}$ and $p_{i,c}$ represent the label and the predicted probability of the $i$-th sample. The supervised contrastive loss $\mathcal{L}_{con}$ is defined as:

\begin{equation}
    \mathcal{L}_{con} = -\sum_{i \in I} \frac{1}{|P(i)|} \sum_{p \in P(i)} \log \frac{\exp \left(z_{i} \cdot z_{p} / \gamma\right)}{\sum_{a \in A(i)} \exp \left(z_{i} \cdot z_{a} / \gamma\right)},
\end{equation}
where $I\equiv\{1 \ldots 2 K\}$ is the batch of contrastive sample pairs, $A(i) \equiv I \backslash\{i\}$ and $i$ is the index of positive sample. $P(i)\equiv {p \in A(i): y_i = y_p}$ is the index set of the positive samples in the batch distinct from $i$, and $y_i$ is the label of the $i$-th sample in the batch. $|P(i)|$ is the number of samples in $P(i)$. $z_{i}$ and $z_{p}$ are the anchor feature and the positive features. $\gamma$ is the temperature parameter. Following~\cite{yang2023co3net}, the weights of cross-entropy and contrastive losses, $w_{ce}$ and $w_{con}$, are set to 0.8 and 0.2, respectively.

The total loss $\mathcal{L}$ is then computed as the weighted sum of the losses above, which is given as follows:
\begin{equation}
    \mathcal{L} = \mathcal{L}_{task} + \mathcal{L}_{prox}^A + \mathcal{L}_{prox}^{g} + \mathcal{L}_{MSE},     
    \label{eq:all}
\end{equation}
where $\mathcal{L}_{prox}^A$ and $\mathcal{L}_{prox}^{g}$ denotes the proximal loss calculated with the anchor model and the global model based on Eq.~\eqref{equ:proximal}.


The main steps of PSFed-Palm are summarized in Algorithm~\ref{alg:PSFed-Palm}. Notably, there is no data sharing during the training phase, rendering PSFed-Palm compatible with distributed settings. A key aspect is that the anchor and global models remain fixed during local training, which ensures that the knowledge encapsulated in these models remains consistent and unaltered during the local training phase.






\subsection{Deployment}

Once the communication rounds are completed, the server distributes the final aggregated model to all clients. These clients then integrate the model into their local verification systems. The PSFed-Palm framework contributes to achieving spectrum consistency within the network, thereby eliminating users' need to register different spectrum palmprint images separately. Instead, clients can employ diverse spectrum acquisition devices for flexible deployment and select the most suitable one based on various practical needs, such as verification time (day or night). PSFed-Palm enables users to verify their identity by matching their different spectrum templates with the registered spectrum template.

In the past, most palmprint verification systems send data to the server for training and send feature templates to the server for verification during deployment. Our method eliminates the need to transmit raw data or templates. Hence, the privacy and security of users' personal information remain intact during the verification and training, as no data is transferred to other parties. Furthermore, transitioning from single spectrum-based verification to cross-spectrum-based verification is smooth and user-friendly, requiring no additional actions from the users. This seamless upgrade empowers users to adopt the new verification method effortlessly, enjoying the benefits of enhanced privacy and improved palmprint verification without disrupting their regular usage patterns.

\section{Evaluation}

\subsection{Settings}
We validated the proposed PSFed-Palm on the public Multi-Spectral dataset~\cite{zhang2009online}. This dataset contains four sub-datasets collected based on the Red, Green, Blue, and NIR spectra. Each sub-dataset was collected in two separate sessions from 500 palms. In each session, each palm acquired six images. Hence, there are 6000 palm images in each sub-dataset.

The only FL-based palm verification method~\cite{shao2020towards} is incompatible with our setting, as it assumes that the server and clients share a set of public data, which may introduce privacy risks. In contrast, our paper strictly adheres to a privacy-preserving setting, where clients exclusively possess the data, and no data transfer is permitted during the training phase. To evaluate the performance of the proposed PSFed-Palm, we compare it with three other methods: "w/o FL", which represents independent training on each client, FedAvg~\cite{mcmahan2017communication}, FedProx~\cite{li2020federated}, FedBN~\cite{li2021fedbn}, and FedPer~\cite{arivazhagan2019federated}. These methods serve as baselines for performance comparison in our experiments.

This paper adopts CompNet \cite{liang2021compnet} as the base model for PSFed-Palm. The simulation is done on the PyTorch framework and optimized by Adam optimizer\cite{kingma2014adam} with a learning rate of 0.01. The batch size, communication round, and local training epoch are set to 512, 100, and 3, respectively. The experimental environment is as follows: AMD Ryzen 7 5800X CPU @3.80GHz, 32GB RAM, and an NVIDIA GTX 3080Ti GPU.

To quantitatively evaluate the performance of different methods, Genuine Acceptance Rate (GAR), False Accept Rate (FAR), and Receiver Operating Characteristic (ROC) curve~\cite{JainBiom} are adopted. The ROC curve is a GAR versus FAR plot that evaluates the trade-off between GAR and FAR. A superior performance is indicated by a ROC curve that closely approaches the top left corner of the plot, indicating a higher GAR and lower FAR. On the other hand, Equal Error Rate (EER) represents the point on the ROC curve where the FAR is equal to the false rejection rate (FRR), where $FRR = 1 - GAR$. Consequently, a lower EER signifies the better performance of the method.

In this paper, the cosine distance is used to calculate the matching distance between two feature vectors $\textbf{v}_i$ and $\textbf{v}_j$ extracted from the trained base model, i.e., $dis(\textbf{v}_i,\textbf{v}_i) =\arccos{(\textbf{v}_i,\textbf{v}_i)}$. A lower cosine distance value indicates a higher similarity between the templates and vice versa.

\subsection{Experiments}

This section presents verification experiments conducted with a few training samples, which utilize two palmprint images per spectrum. This setup enables us to assess the efficacy of our proposed method when faced with limited training data, which is consistent with the real scenario wherein a sparse number of samples are accessible for each spectrum.

\begin{figure}
    \centering
    \includegraphics[width=\textwidth]{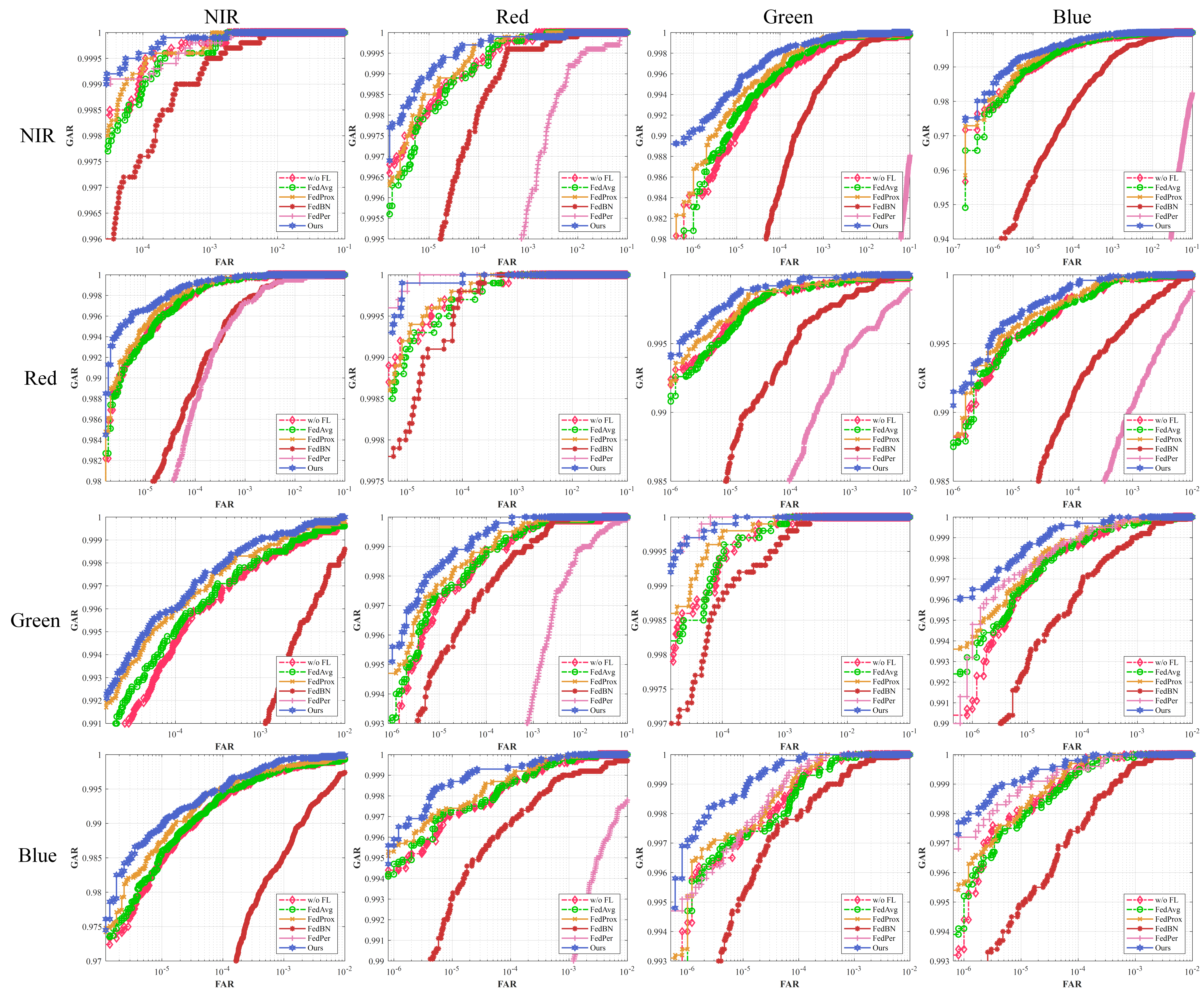}
    \caption{The ROC curves under different spectrums pairing.}
    \label{fig: Few_Roc}
\end{figure}

We adopt a specific spectrum to train the network and employ images from other spectrums for evaluation. Fig. \ref{fig: Few_Roc} presents the ROC curves under different spectrum pairing configurations. It can be seen that our PSFed-Palm consistently demonstrates superior performance over other approaches across most scenarios. FedBN and FedPer are two personalized FL methods that aim to get personalized models for different models. Hence, the two methods only consider the performance in the local data, which means the two methods would give up the spectrum consistency and overly focus on the performances in the local spectra data. Unlike the personalized FL methods, we observe that FL-based methods outperform non-Federated Learning ('w/o FL'), which supports the advantages of incorporating FL techniques in palmprint biometrics. 

\begin{table}[]
\caption{EERs under different spectrums pairing for 'w/o FL' (\%).}
\small
\begin{tabular}{@{}ccccc|c@{}}
\toprule
  & NIR  & Red & Green & Blue & Average\\
\midrule
NIR     & 0.03000 & 0.03000   & 0.12000     & 0.13078 & 0.07769 \\
Red     & 0.05000 & 0.02000   & 0.06000     & 0.04092  & 0.04273 \\
Green   & 0.22244 & 0.05922   & 0.03000     & 0.04529 & 0.08924 \\
Blue    & 0.15659 & 0.04383   & 0.02423 & 0.01359 & 0.05956 \\ 
\midrule
Average & 0.11476 & 0.03826 & 0.05856 & 0.05765 & 0.06731  \\    
\botrule
\end{tabular}
\label{tab:fewshot/wo}
\end{table}

\begin{table}[]
\caption{EERs under different spectrums pairing for FedAvg (\%).}
\small
\begin{tabular}{@{}ccccc|c@{}}
\toprule
  & NIR  & Red & Green & Blue & Average\\
\midrule
NIR     & 0.04000 & 0.02445 & 0.10707 & 0.10000      & 0.06788 \\
Red     & 0.05000 & 0.02000     & 0.06000     & 0.04491  & 0.04373 \\
Green   & 0.15000 & 0.05257 & 0.03000     & 0.04000     & 0.06814 \\
Blue    & 0.16000 & 0.05000     & 0.03705 & 0.02114 & 0.06705 \\
\midrule
Average & 0.10000  & 0.03675 & 0.05853 & 0.05151 & 0.06170  \\
\botrule
\end{tabular}
\label{tab:fewshot/fedavg}
\end{table}

\begin{table}[]
\caption{EERs under different spectrums pairing for FedProx (\%).}
\small
\begin{tabular}{@{}ccccc|c@{}}
\toprule
  & NIR  & Red & Green & Blue & Average\\
\midrule
NIR     & 0.04000     & 0.02000     & 0.08000     & 0.09000     & 0.05750   \\
Red     & 0.03735 & 0.01413 & 0.05000     & 0.04000     & 0.03537 \\
Green   & 0.12727 & 0.04194 & 0.02000     & 0.04000     & 0.05730 \\
Blue    & 0.15000     & 0.03561 & 0.02277 & 0.01449 & 0.05572 \\ 
\midrule
Average & 0.08866 & 0.02792 & 0.04319 & 0.04612 & 0.05147 \\
\botrule
\end{tabular}
\label{tab:fewshot/fedprox}
\end{table}

\begin{table}[]
\caption{EERs under different spectrums pairing for FedBN (\%).}
\small
\begin{tabular}{@{}ccccc|c@{}}
\toprule
  & NIR  & Red & Green & Blue & Average\\
\midrule
NIR     & 0.07611 & 0.04000 & 0.25072     & 0.36000     & 0.18171   \\
Red     & 0.19000 & 0.01936 & 0.12930     & 0.02000     & 0.13466 \\
Green   & 0.40122 & 0.09074 & 0.05776     & 0.09930     & 0.16226 \\
Blue    & 0.52567 & 0.10000 & 0.07000     & 0.07000     & 0.19142 \\ 
\midrule
Average & 0.29825 & 0.06253 & 0.12694 & 0.18233 & 0.16751 \\
\botrule
\end{tabular}
\label{tab:fewshot/fedbn}
\end{table}

\begin{table}[]
\caption{EERs under different spectrums pairing for FedPer (\%).}
\small
\begin{tabular}{@{}ccccc|c@{}}
\toprule
  & NIR  & Red & Green & Blue & Average\\
\midrule
NIR     & 0.04000 & 0.23479 & 3.42000     & 4.25000     & 1.98620   \\
Red     & 0.18866 & 0.00168 & 0.32128     & 0.41188     & 0.23088 \\
Green   & 5.30000 & 0.28062 & 0.00645     & 0.03988     & 1.40674 \\
Blue    & 7.49319 & 0.41816 & 0.02000     & 0.02752     & 1.98971 \\ 
\midrule
Average & 3.25546 & 0.23381 & 0.94193 & 1.18232 & 1.40338 \\
\botrule
\end{tabular}
\label{tab:fewshot/fedper}
\end{table}

\begin{table}[]
\caption{EERs under different spectrums for PSFed-Palm (\%).}
\small
\begin{tabular}{@{}ccccc|c@{}}
\toprule
  & NIR  & Red & Green & Blue & Average\\
\midrule
NIR     & 0.02000    & 0.01768 & 0.08000  & 0.06661 & \textbf{0.04607} \\
Red     & 0.03361 & 0.01000    & 0.02000  & 0.03000 & \textbf{0.02340}  \\
Green   & 0.10000 & 0.02000 & 0.01000 & 0.02429 & \textbf{0.03857} \\
Blue    & 0.08363 & 0.03000 & 0.01014 & 0.01309 & \textbf{0.03421} \\ \midrule
Average & \textbf{0.05931} & \textbf{0.01942} & \textbf{0.03004} & \textbf{0.03350}  & \textbf{0.03556}\\
\botrule
\end{tabular}
\label{tab:fewshot/ours}
\end{table}

The results have also been presented in EER, as depicted in Tabs.~\ref{tab:fewshot/wo} to~\ref{tab:fewshot/ours}. From Tab.~\ref{tab:fewshot/wo}, we observe that the performance supports our assumption, demonstrating that similar spectrum images tend to exhibit higher degrees of similarity. For instance, in the case of NIR spectrum images, the EERs for the corresponding images from the Green or Blue spectrums are significantly higher than the ones from the Red spectrum. This trend is consistently observed across all spectra, substantiating our assumption's validity and reinforcing our proposed approach's underlying rationale.

Furthermore, we observe that the performance of FedProx surpasses FedAvg, which shows the effectiveness of incorporating the proximal item in FL training. In Tabs.~\ref{tab:fewshot/wo} to~\ref{tab:fewshot/ours}, it is evident that our proposed PSFed-Palm consistently achieves the lowest EER across all scenarios compared to other methods. The personalized FL methods can perform satisfactorily if the image pairs are from the same spectrum. However, they would fail to match the image pairs from different spectrums. This outcome serves as evidence of the effectiveness of our spectrum-consistency strategy.

\begin{figure}
    \centering
    \includegraphics[width=\textwidth]{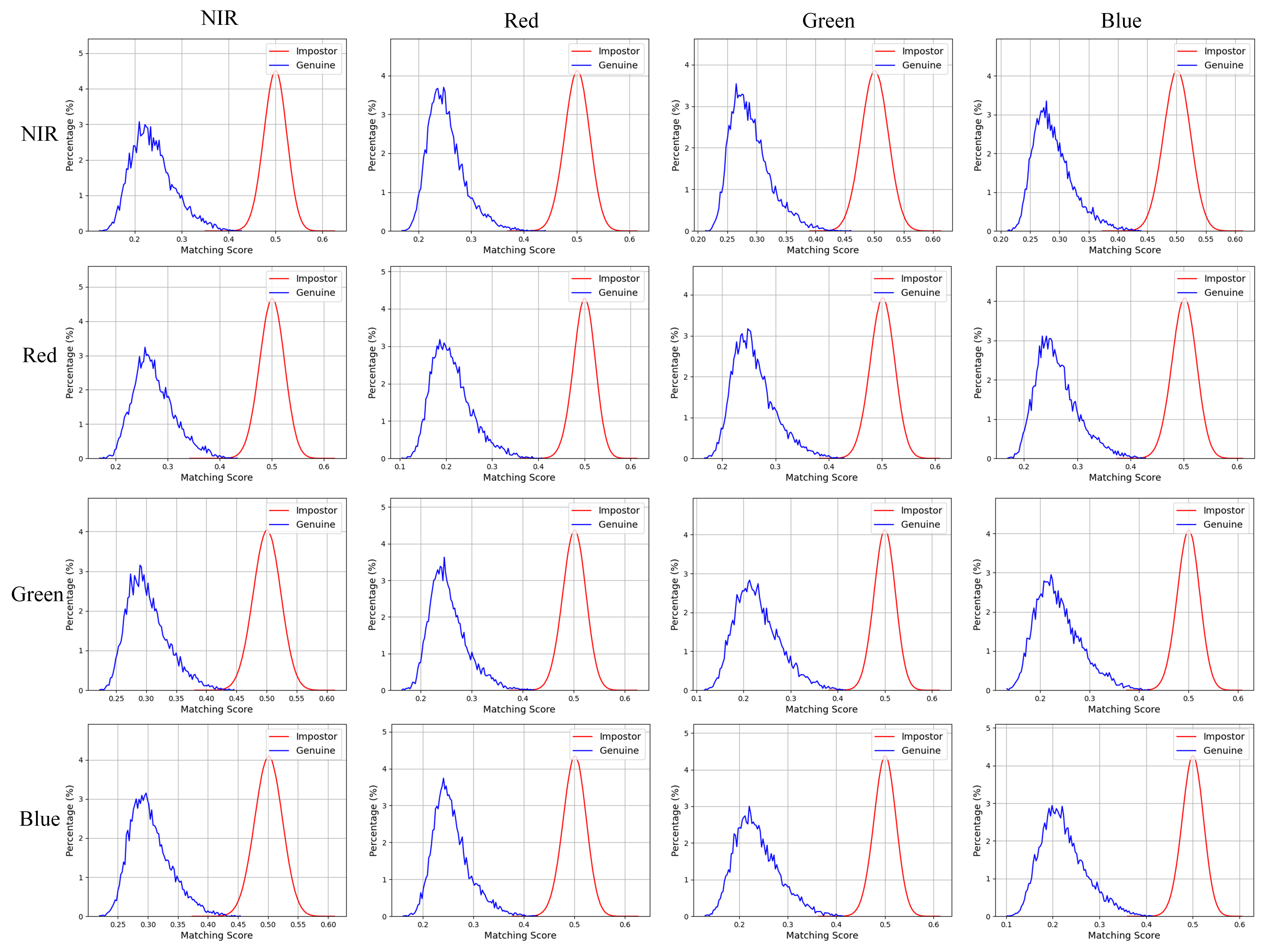}
    \caption{The genuine- and imposter-matching distributions under different spectrums pairing.}
    \label{fig: Few_GI}
\end{figure}

Fig. \ref{fig: Few_GI} shows the genuine- and imposter-matching distributions. Notably, a distinct interval is observed between the genuine- and imposter-matching distributions, with relatively small overlapping ranges. Additionally, our method consistently demonstrates these distinguishing properties across all scenarios. This observation reinforces the efficacy of our approach in achieving a clear separation between genuine and imposter palmprint matches, contributing to improved verification accuracy and consistency.

\subsection{Ablation Studies}

\subsubsection{The number of training samples}

\begin{figure}
    \centering
    \includegraphics[width=.7\textwidth]{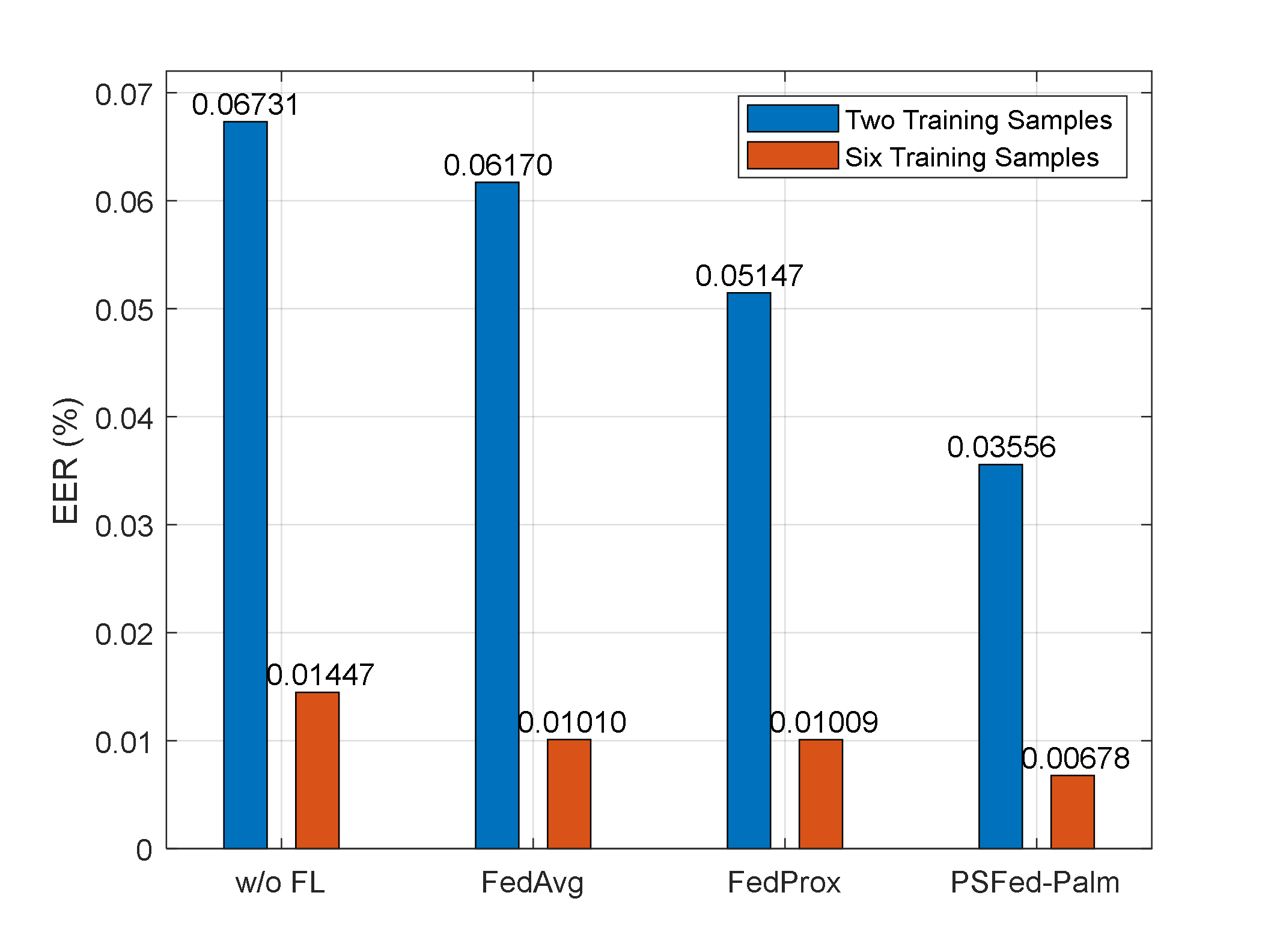}
    \caption{The comparison results of different methods under different training samples size.}
    \label{fig:Train_Sample}
\end{figure}

In this subsection, we investigate the impact of training set size on the PSFed-Palm. In the previous section, we utilized a small number of training samples, with only two palmprint images per spectrum. We have conducted an additional experiment using six palmprint images per spectrum for training. The results were derived from the average verification results of all spectrum pairs as shown in Fig.~\ref{fig:Train_Sample}. We should note that the EERs of FedBN and FedPer are 0.3819 and 0.8200, respectively. The two methods are obviously worse than other methods, so the comparisons of them with different training samples were not illustrated in Fig.~\ref{fig:Train_Sample}.

Despite the apparent narrowing of the advantage gap observed in Fig.~\ref{fig:Train_Sample} when utilizing six training samples, our method exhibits a significant superiority over competing approaches in both cases. Notably, during training with only two samples, our method displayed a remarkable performance improvement of 47.2\% over 'w/o FL', 42.4\% over 'FedAvg', and 30.9\% over 'FedProx', respectively. These performance improvements became even more evident when employing six training samples, up to 53.14\% over 'w/o FL', 32.9\% over 'FedAvg', and 32.8\% over 'FedProx'.

PSFed-Palm demonstrates consistent and promising performance even in this setting with an increased training sample size. We attribute this preservation of performance to the different types of knowledge learned by the global and anchor models, namely, global knowledge and spectrum group knowledge. Utilizing these different forms of knowledge enables the adjustment of local model optimization directions, effectively mitigating overfitting to local data and spectrum specifics. Consequently, the training process becomes more stable and leads to satisfactory performance.

\subsubsection{Loss Functions}

We evaluate the effectiveness of each loss component in PSFed-Palm, and the results are listed in Tab.~\ref{tab:abla}. In Tab.~\ref{tab:abla}, it became evident that relying solely on anchor models to constrain local models using a parameter or feature representation led to unsatisfactory results. This was mainly attributed to model drift, a phenomenon where local models gradually deviate from the global model during training.

However, a significant performance improvement was observed when we introduced the global proximal loss in combination with the spectrum-consistency loss (Anchor Proximal Loss + MSE loss). This combination effectively addressed the model drift issue and improved overall performance. Moreover, when all three proposed loss components were combined, the overall performance was further boosted. The results demonstrated the effectiveness of each loss component, and integrating these losses also proved highly effective in improving the performance of PSFed-Palm.

\begin{table*}[]
\centering
\caption{Ablation of Different Components in the Total Loss Function.}
\label{tab:abla}
\begin{tabular}{ccccc}
\hline
\centering
Global Proximal Loss  & Anchor Proximal Loss  & MSE Loss & EER
                \\ \hline
\ding{51} & \ding{55} & \ding{55} & 0.01009\% \\ 
\ding{55} & \ding{51} & \ding{55} & 0.01104\% \\    
\ding{55} & \ding{55} & \ding{51} & 0.01352\% \\    
\ding{55} & \ding{51} & \ding{51} & 0.01190\% \\ 
\ding{51} & \ding{51} & \ding{55} & 0.00886\% \\
\ding{51} & \ding{55} & \ding{51} & 0.00828\% \\
\ding{51} & \ding{51} & \ding{51} & \textbf{0.00678\%} \\ \hline
\end{tabular}
\end{table*}

\subsubsection{Hyperparameters Tuning}
\begin{figure}[!t]
\vspace{8pt}
	\centering
	\begin{minipage}[t]{0.48\textwidth}
	\includegraphics[width=\textwidth]{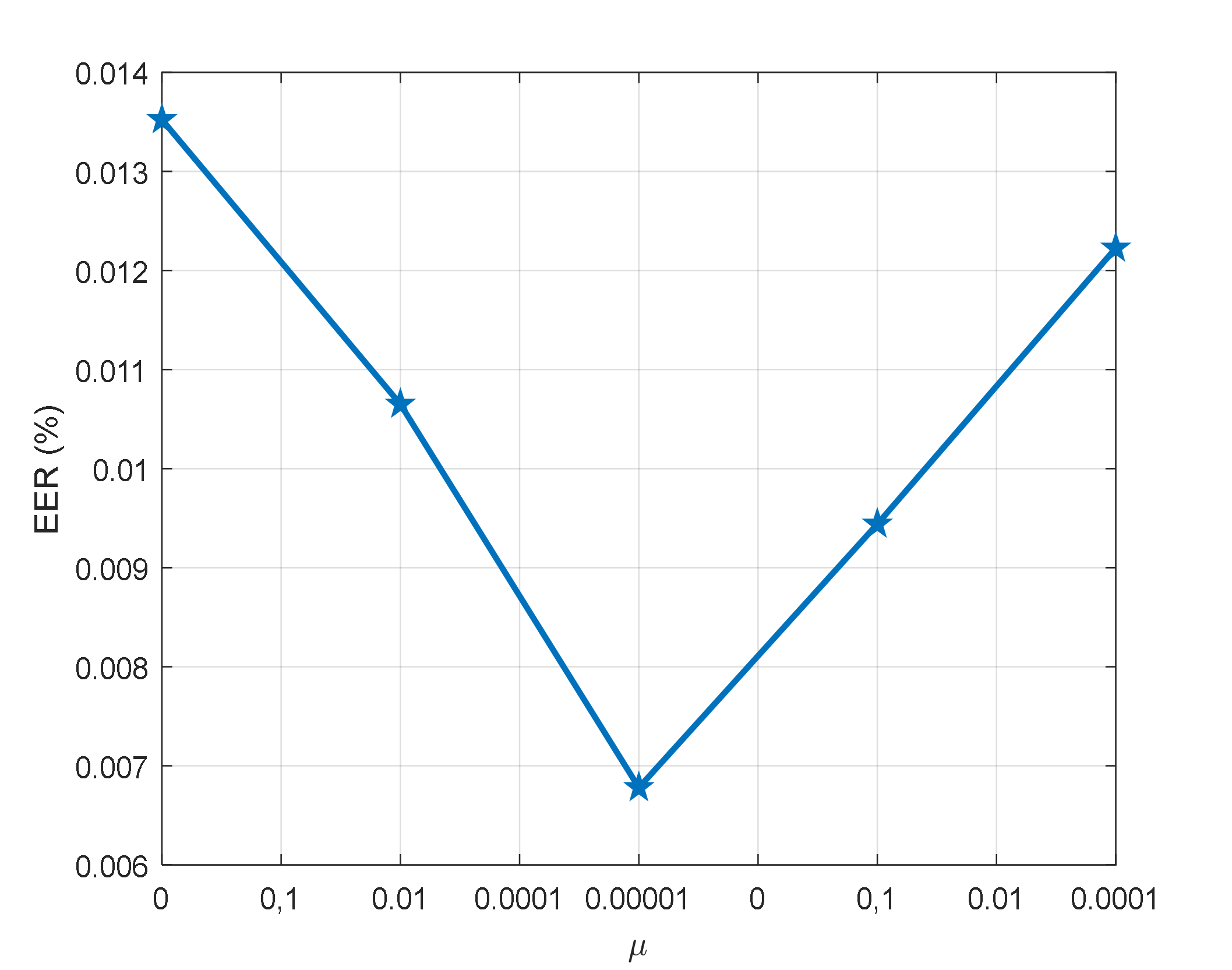}
	\centerline{(a)}
	\end{minipage}
	\begin{minipage}[t]{0.48\textwidth}
	\includegraphics[width=\textwidth]{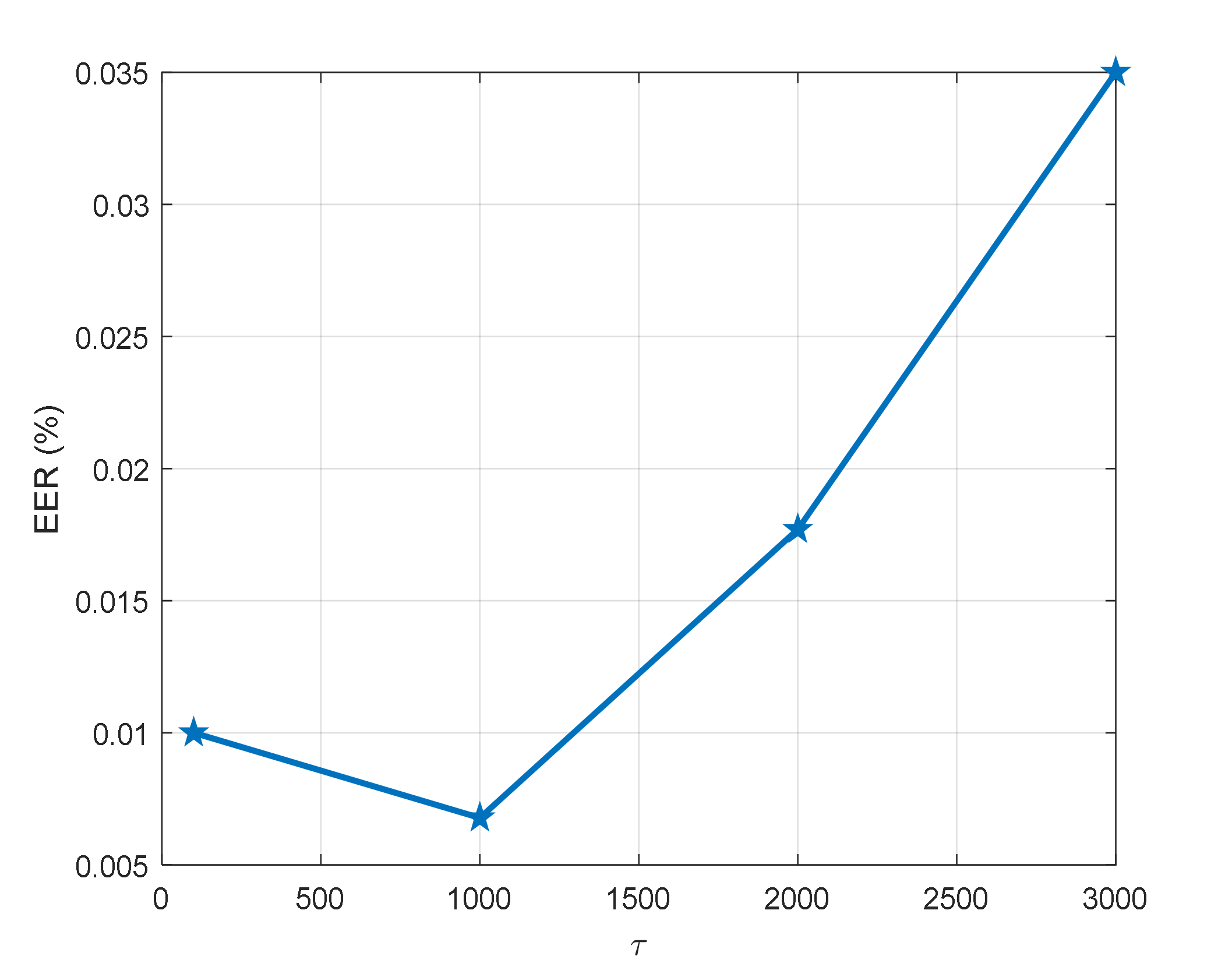}
	\centerline{(b)}
	\end{minipage}
 \vspace{5pt}
\caption{The ablation studies of hyperparameters. (a)-(b) represent $\mu$ and $\tau$ in Eqs.~\eqref{equ:proximal} and~\eqref{eq:all}, respectively.}
\label{fig: para}
\end{figure}

Our proposed method has two hyperparameters: $\mu$ and $\tau$ in Eqs. \eqref{equ:proximal} and \eqref{eq:all}. We attempt to investigate the impacts of these hyperparameters on the training process. The results presented in Fig. \ref{fig: para} suggest that if the values of these hyperparameters are too small, the corresponding losses would not effectively constrain the training process. Conversely, if they are too large, the network may prioritize the proximal losses and overlook the task loss, resulting in suboptimal performance. 

Based on the experiments, we empirically set $\mu$ to 0.01 and $\tau$ to 1000 to balance practical constraints and task performance. It is important to note that the other experiments in this paper follow the same hyperparameter setting.

\subsubsection{Local Training Rounds}
\begin{figure}[]
    \centerline{\includegraphics[width=.7\textwidth]{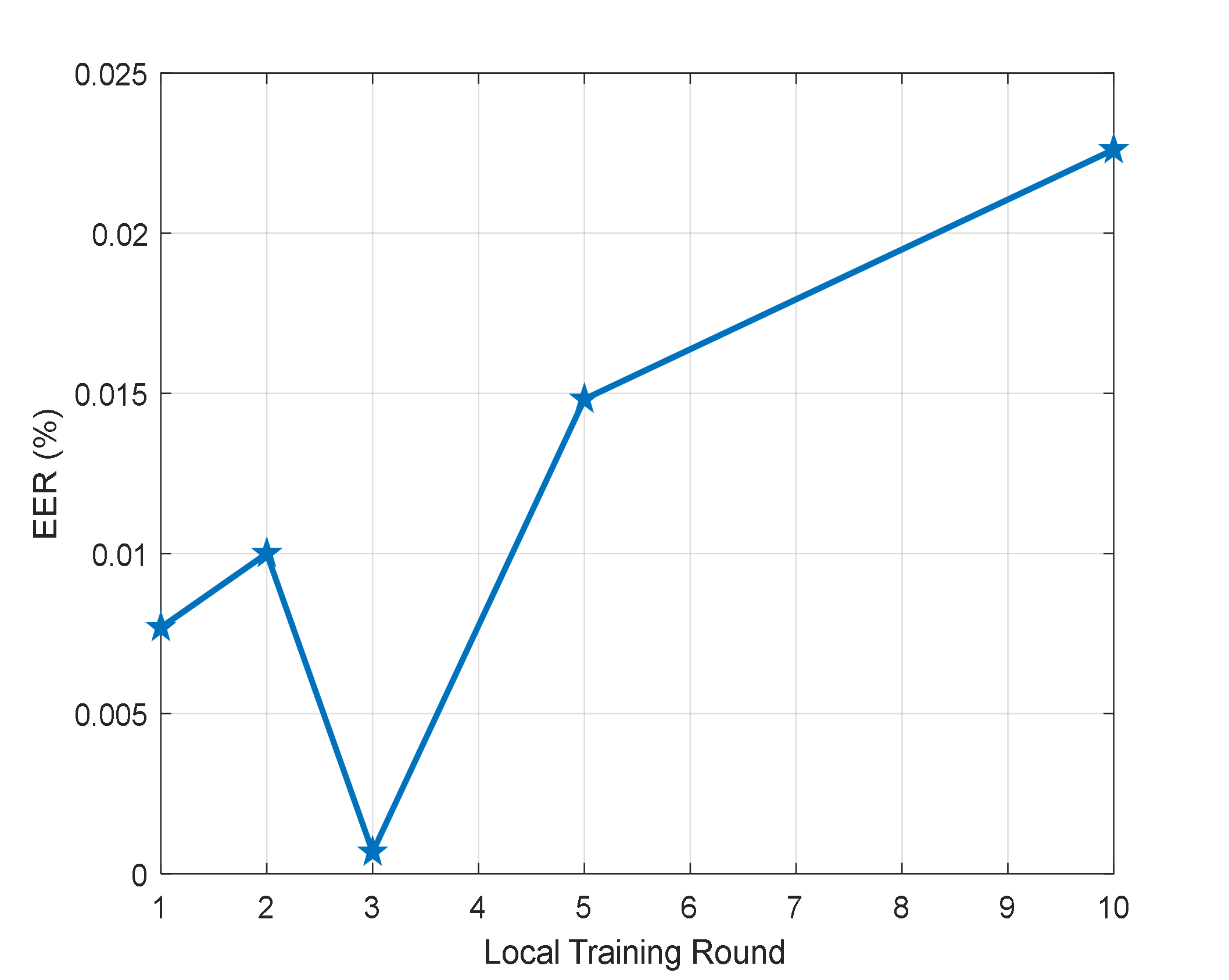}}
    \caption{The relationship between the performance and the number of local training rounds.}
    \label{fig:communication}
\end{figure}

This ablation explores the relationship between verification performance and the number of local training rounds. To ensure a fair comparison, we set a fixed number of local training rounds as $E \times R = 300$. The results are shown in Fig. \ref{fig:communication}. It can be noticed that when the number of local iterations is small, the difference in performance is insignificant. However, as the number of local training rounds increases, the performance declines noticeably. This outcome lies in the heterogeneity of data across different clients and the excessive number of local iterations, which aggravate the problem of model drift and significantly affect the overall effectiveness of the trained model. We recommend avoiding an excessively high number of local training rounds to mitigate the adverse effects of model drift and maintain the overall performance of the proposed method.

\subsubsection{The framework's compatibility}
This subsection assesses the PSFed-Palm framework's compatibility, utilizing CO3Net~\cite{yang2023co3net} as the underlying base model. The results are given in Tab.~\ref{tab:co3}. Notably, CO3Net contains significantly larger parameters than CompNet, highlighting its heightened susceptibility to the heterogeneity problem. Consequently, even minor optimization errors can lead to a decline in performance. The disparities in performance between FedAvg and PSFed-Palm are more pronounced than those observed in prior experiments with CompNet as the base model. Additionally, the extensive number of parameters in the personalized fully connected layers of FedPer has a negative impact on its performance, as these layers do not participate in the aggregation process. As a result, the network becomes excessively fixated on local spectra, leading to overall performance deterioration. Our experimental results demonstrate the compatibility of the proposed method, confirming its capability to maintain superior performance across different base models.

\begin{table*}[]
\centering
\caption{Comparison results with CO3Net as the base model.}
\label{tab:co3}
\begin{tabular}{ccccccc}
\hline
\centering
  & w/o FL & FedAvg & FedProx & FedBN & FedPer & PSFed-Palm
                \\ \hline
EER & 0.01912\% & 0.01407\% & 0.01079\% & 0.15851\% & 1.69447\% & \textbf{0.00868\%} \\ \hline
\end{tabular}
\end{table*}

\subsection{The framework's generality}

Finally, we conduct experiments using the CASIA Multi-Spectral dataset~\cite{hao2008multispectral}, a publicly available dataset containing 7,200 palmprint images from 200 palms. These palmprint images were captured using six spectrums: 460nm, 630nm, 700nm, 850nm, 940nm, and white light. For our experimental setup, we deliberately select the 460nm, 630nm, 850nm, and 940nm spectrums to create a more challenging non-iid scenario. Specifically, two palmprint images per spectrum are utilized as training samples, and CompNet is used as the base model.

The experimental results are presented in Tab.~\ref{tab:CASIA}. It is evident from the results that the performance of personalized FL is inferior. This discrepancy can be attributed to the difference in optimization objectives and the emphasis on spectrum consistency between personalized FL and PSFed-Palm. Personalized FL methods solely concentrate on enhancing local performance, often disregarding the performance in other spectrum data. Consequently, personalized FL tends to be in opposition to spectrum consistency. On the other hand, our method outperforms other approaches, underscoring the robustness and generalizability of the proposed PSFed-Palm framework.

\begin{table*}[]
\centering
\caption{Comparison results in CASIA Multi-Spectral dataset.}
\label{tab:CASIA}
\begin{tabular}{ccccccc}
\hline
\centering
  & w/o FL & FedAvg & FedProx & FedBN & FedPer & PSFed-Palm
                \\ \hline
EER & 2.01202\% & 1.60464\% & 1.61989\% & 2.26141\% & 6.43788\% & \textbf{1.22485\%} \\ \hline
\end{tabular}
\end{table*}

\section{Conclusions}

This paper introduces a novel physics-driven spectrum-consistent federated learning approach for palmprint verification, ensuring privacy by eliminating the need for local data transfer during the training phase. The clients are partitioned into two groups based on their spectrum types, with corresponding anchor models designed accordingly. This approach guarantees spectrum consistency and effectively bridges the gap between different spectrum templates, enhancing verification accuracy and robustness. The experimental results validate the effectiveness and robustness of the proposed method. It is important to note that this work assumes the participation of all spectrums during the training process. However, exploring strategies to achieve spectrum consistency without specific spectrums poses an exciting and challenging avenue for future research.

\bmhead{Acknowledgments}
This work was supported in part by the National Natural Science Foundation of China under Grants 62271335; in part by the Sichuan Science and Technology Program under Grant 2021JDJQ0024; and in part by the Sichuan University “From 0 to 1” Innovative Research Program under Grant 2022SCUH0016.

\bmhead{Data availability statement}
The data that support the findings of this study are available from PolyU~\cite{zhang2009online} and CASIA~\cite{hao2008multispectral} but restrictions apply to the availability of these data, which were used under license for the current study, and so are not publicly available. Data are however available from the authors upon reasonable request and with permission of PolyU~\cite{zhang2009online} and CASIA~\cite{hao2008multispectral}.
 
\bibliography{ref}


\begin{thebibliography}{47}
\ifx \bisbn   \undefined \def \bisbn  #1{ISBN #1}\fi
\ifx \binits  \undefined \def \binits#1{#1}\fi
\ifx \bauthor  \undefined \def \bauthor#1{#1}\fi
\ifx \batitle  \undefined \def \batitle#1{#1}\fi
\ifx \bjtitle  \undefined \def \bjtitle#1{#1}\fi
\ifx \bvolume  \undefined \def \bvolume#1{\textbf{#1}}\fi
\ifx \byear  \undefined \def \byear#1{#1}\fi
\ifx \bissue  \undefined \def \bissue#1{#1}\fi
\ifx \bfpage  \undefined \def \bfpage#1{#1}\fi
\ifx \blpage  \undefined \def \blpage #1{#1}\fi
\ifx \burl  \undefined \def \burl#1{\textsf{#1}}\fi
\ifx \doiurl  \undefined \def \doiurl#1{\url{https://doi.org/#1}}\fi
\ifx \betal  \undefined \def \betal{\textit{et al.}}\fi
\ifx \binstitute  \undefined \def \binstitute#1{#1}\fi
\ifx \binstitutionaled  \undefined \def \binstitutionaled#1{#1}\fi
\ifx \bctitle  \undefined \def \bctitle#1{#1}\fi
\ifx \beditor  \undefined \def \beditor#1{#1}\fi
\ifx \bpublisher  \undefined \def \bpublisher#1{#1}\fi
\ifx \bbtitle  \undefined \def \bbtitle#1{#1}\fi
\ifx \bedition  \undefined \def \bedition#1{#1}\fi
\ifx \bseriesno  \undefined \def \bseriesno#1{#1}\fi
\ifx \blocation  \undefined \def \blocation#1{#1}\fi
\ifx \bsertitle  \undefined \def \bsertitle#1{#1}\fi
\ifx \bsnm \undefined \def \bsnm#1{#1}\fi
\ifx \bsuffix \undefined \def \bsuffix#1{#1}\fi
\ifx \bparticle \undefined \def \bparticle#1{#1}\fi
\ifx \barticle \undefined \def \barticle#1{#1}\fi
\bibcommenthead
\ifx \bconfdate \undefined \def \bconfdate #1{#1}\fi
\ifx \botherref \undefined \def \botherref #1{#1}\fi
\ifx \url \undefined \def \url#1{\textsf{#1}}\fi
\ifx \bchapter \undefined \def \bchapter#1{#1}\fi
\ifx \bbook \undefined \def \bbook#1{#1}\fi
\ifx \bcomment \undefined \def \bcomment#1{#1}\fi
\ifx \oauthor \undefined \def \oauthor#1{#1}\fi
\ifx \citeauthoryear \undefined \def \citeauthoryear#1{#1}\fi
\ifx \endbibitem  \undefined \def \endbibitem {}\fi
\ifx \bconflocation  \undefined \def \bconflocation#1{#1}\fi
\ifx \arxivurl  \undefined \def \arxivurl#1{\textsf{#1}}\fi
\csname PreBibitemsHook\endcsname

\bibitem[\protect\citeauthoryear{Han et~al.}{2022}]{han2022personalized}
\begin{barticle}
\bauthor{\bsnm{Han}, \binits{C.}},
\bauthor{\bsnm{Shan}, \binits{S.}},
\bauthor{\bsnm{Kan}, \binits{M.}},
\bauthor{\bsnm{Wu}, \binits{S.}},
\bauthor{\bsnm{Chen}, \binits{X.}}:
\batitle{Personalized convolution for face recognition}.
\bjtitle{Int. J. Comput. Vis.}
\bvolume{130}(\bissue{2}),
\bfpage{344}--\blpage{362}
(\byear{2022})
\end{barticle}
\endbibitem

\bibitem[\protect\citeauthoryear{Gomez-Barrero
  et~al.}{2021}]{gomez2021biometrics}
\begin{botherref}
\oauthor{\bsnm{Gomez-Barrero}, \binits{M.}},
\oauthor{\bsnm{Drozdowski}, \binits{P.}},
\oauthor{\bsnm{Rathgeb}, \binits{C.}},
\oauthor{\bsnm{Patino}, \binits{J.}},
\oauthor{\bsnm{Todisco}, \binits{M.}},
\oauthor{\bsnm{Nautsch}, \binits{A.}},
\oauthor{\bsnm{Damer}, \binits{N.}},
\oauthor{\bsnm{Priesnitz}, \binits{J.}},
\oauthor{\bsnm{Evans}, \binits{N.}},
\oauthor{\bsnm{Busch}, \binits{C.}}:
Biometrics in the era of covid-19: challenges and opportunities.
arXiv preprint arXiv:2102.09258
(2021)
\end{botherref}
\endbibitem

\bibitem[\protect\citeauthoryear{Zhang et~al.}{2003}]{zhang2003online}
\begin{barticle}
\bauthor{\bsnm{Zhang}, \binits{D.}},
\bauthor{\bsnm{Kong}, \binits{W.-K.}},
\bauthor{\bsnm{You}, \binits{J.}},
\bauthor{\bsnm{Wong}, \binits{M.}}:
\batitle{Online palmprint identification}.
\bjtitle{IEEE Trans. Pattern Anal. Mach. Intell.}
\bvolume{25}(\bissue{9}),
\bfpage{1041}--\blpage{1050}
(\byear{2003})
\end{barticle}
\endbibitem

\bibitem[\protect\citeauthoryear{Xu et~al.}{2018}]{xu2018drcc}
\begin{barticle}
\bauthor{\bsnm{Xu}, \binits{Y.}},
\bauthor{\bsnm{Fei}, \binits{L.}},
\bauthor{\bsnm{Wen}, \binits{J.}},
\bauthor{\bsnm{Zhang}, \binits{D.}}:
\batitle{Discriminative and robust competitive code for palmprint recognition}.
\bjtitle{IEEE Trans. Syst. Man Cybern. Syst.}
\bvolume{48}(\bissue{2}),
\bfpage{232}--\blpage{241}
(\byear{2018})
\end{barticle}
\endbibitem

\bibitem[\protect\citeauthoryear{Liang et~al.}{2021}]{liang2021compnet}
\begin{barticle}
\bauthor{\bsnm{Liang}, \binits{X.}},
\bauthor{\bsnm{Yang}, \binits{J.}},
\bauthor{\bsnm{Lu}, \binits{G.}},
\bauthor{\bsnm{Zhang}, \binits{D.}}:
\batitle{Compnet: Competitive neural network for palmprint recognition using
  learnable gabor kernels}.
\bjtitle{IEEE Signal Process. Lett.}
\bvolume{28},
\bfpage{1739}--\blpage{1743}
(\byear{2021})
\end{barticle}
\endbibitem

\bibitem[\protect\citeauthoryear{Yang et~al.}{2023}]{yang2023co3net}
\begin{barticle}
\bauthor{\bsnm{Yang}, \binits{Z.}},
\bauthor{\bsnm{Xia}, \binits{W.}},
\bauthor{\bsnm{Qiao}, \binits{Y.}},
\bauthor{\bsnm{Lu}, \binits{Z.}},
\bauthor{\bsnm{Zhang}, \binits{B.}},
\bauthor{\bsnm{Leng}, \binits{L.}},
\bauthor{\bsnm{Zhang}, \binits{Y.}}:
\batitle{Co3net: Coordinate-aware contrastive competitive neural network for
  palmprint recognition}.
\bjtitle{IEEE Trans. Instrum. Meas.}
\bvolume{72},
\bfpage{3276506}
(\byear{2023})
\end{barticle}
\endbibitem

\bibitem[\protect\citeauthoryear{Zhang et~al.}{2010}]{zhang2009online}
\begin{barticle}
\bauthor{\bsnm{Zhang}, \binits{D.}},
\bauthor{\bsnm{Guo}, \binits{Z.}},
\bauthor{\bsnm{Lu}, \binits{G.}},
\bauthor{\bsnm{Zhang}, \binits{L.}},
\bauthor{\bsnm{Zuo}, \binits{W.}}:
\batitle{An online system of multispectral palmprint verification}.
\bjtitle{IEEE Trans. Instrum. Meas.}
\bvolume{59}(\bissue{2}),
\bfpage{480}--\blpage{490}
(\byear{2010})
\end{barticle}
\endbibitem

\bibitem[\protect\citeauthoryear{Dong et~al.}{2022}]{dong2022co}
\begin{barticle}
\bauthor{\bsnm{Dong}, \binits{X.}},
\bauthor{\bsnm{Khan}, \binits{M.K.}},
\bauthor{\bsnm{Leng}, \binits{L.}},
\bauthor{\bsnm{Teoh}, \binits{A.B.J.}}:
\batitle{Co-learning to hash palm biometrics for flexible iot deployment}.
\bjtitle{IEEE Internet of Things Journal}
\bvolume{9}(\bissue{23}),
\bfpage{23786}--\blpage{23794}
(\byear{2022})
\end{barticle}
\endbibitem

\bibitem[\protect\citeauthoryear{Fei et~al.}{2020}]{fei2020feature}
\begin{barticle}
\bauthor{\bsnm{Fei}, \binits{L.}},
\bauthor{\bsnm{Zhang}, \binits{B.}},
\bauthor{\bsnm{Jia}, \binits{W.}},
\bauthor{\bsnm{Wen}, \binits{J.}},
\bauthor{\bsnm{Zhang}, \binits{D.}}:
\batitle{Feature extraction for 3-d palmprint recognition: A survey}.
\bjtitle{IEEE Trans. Instrum. Meas.}
\bvolume{69}(\bissue{3}),
\bfpage{645}--\blpage{656}
(\byear{2020})
\end{barticle}
\endbibitem

\bibitem[\protect\citeauthoryear{Zhong et~al.}{2019}]{zhong2019decade}
\begin{barticle}
\bauthor{\bsnm{Zhong}, \binits{D.}},
\bauthor{\bsnm{Du}, \binits{X.}},
\bauthor{\bsnm{Zhong}, \binits{K.}}:
\batitle{Decade progress of palmprint recognition: A brief survey}.
\bjtitle{Neurocomputing}
\bvolume{328},
\bfpage{16}--\blpage{28}
(\byear{2019})
\end{barticle}
\endbibitem

\bibitem[\protect\citeauthoryear{Zhang et~al.}{2018}]{zhang2018combining}
\begin{barticle}
\bauthor{\bsnm{Zhang}, \binits{S.}},
\bauthor{\bsnm{Wang}, \binits{H.}},
\bauthor{\bsnm{Huang}, \binits{W.}},
\bauthor{\bsnm{Zhang}, \binits{C.}}:
\batitle{Combining modified lbp and weighted src for palmprint recognition}.
\bjtitle{Signal, Image and Video Processing}
\bvolume{12},
\bfpage{1035}--\blpage{1042}
(\byear{2018})
\end{barticle}
\endbibitem

\bibitem[\protect\citeauthoryear{Guo et~al.}{2009}]{guo2009palmprint}
\begin{barticle}
\bauthor{\bsnm{Guo}, \binits{Z.}},
\bauthor{\bsnm{Zhang}, \binits{D.}},
\bauthor{\bsnm{Zhang}, \binits{L.}},
\bauthor{\bsnm{Zuo}, \binits{W.}}:
\batitle{Palmprint verification using binary orientation co-occurrence vector}.
\bjtitle{Pattern Recognit. Lett.}
\bvolume{30}(\bissue{13}),
\bfpage{1219}--\blpage{1227}
(\byear{2009})
\end{barticle}
\endbibitem

\bibitem[\protect\citeauthoryear{Yang et~al.}{2023}]{yang20212TCC}
\begin{barticle}
\bauthor{\bsnm{Yang}, \binits{Z.}},
\bauthor{\bsnm{Leng}, \binits{L.}},
\bauthor{\bsnm{Wu}, \binits{T.}},
\bauthor{\bsnm{Li}, \binits{M.}},
\bauthor{\bsnm{Chu}, \binits{J.}}:
\batitle{Multi-order texture features for palmprint recognition}.
\bjtitle{Artif. Intell. Rev.}
\bvolume{56}(\bissue{2}),
\bfpage{995}--\blpage{1011}
(\byear{2023})
\end{barticle}
\endbibitem

\bibitem[\protect\citeauthoryear{Zhang et~al.}{2012}]{zhang2012fragile}
\begin{barticle}
\bauthor{\bsnm{Zhang}, \binits{L.}},
\bauthor{\bsnm{Li}, \binits{H.}},
\bauthor{\bsnm{Niu}, \binits{J.}}:
\batitle{Fragile bits in palmprint recognition}.
\bjtitle{IEEE Signal Process. Lett.}
\bvolume{19}(\bissue{10}),
\bfpage{663}--\blpage{666}
(\byear{2012})
\end{barticle}
\endbibitem

\bibitem[\protect\citeauthoryear{Sun et~al.}{2005}]{sun2005ordinal}
\begin{bchapter}
\bauthor{\bsnm{Sun}, \binits{Z.}},
\bauthor{\bsnm{Tan}, \binits{T.}},
\bauthor{\bsnm{Wang}, \binits{Y.}},
\bauthor{\bsnm{Li}, \binits{S.Z.}}:
\bctitle{Ordinal palmprint represention for personal identification}.
In: \bbtitle{Proc. IEEE/CVF Int. Conf. Comput. Vis. Pattern Recognit. (CVPR)},
vol. \bseriesno{1},
pp. \bfpage{279}--\blpage{284}
(\byear{2005})
\end{bchapter}
\endbibitem

\bibitem[\protect\citeauthoryear{Yang et~al.}{2021}]{yang2021extreme}
\begin{barticle}
\bauthor{\bsnm{Yang}, \binits{Z.}},
\bauthor{\bsnm{Leng}, \binits{L.}},
\bauthor{\bsnm{Min}, \binits{W.}}:
\batitle{Extreme downsampling and joint feature for coding-based palmprint
  recognition}.
\bjtitle{IEEE Trans. Instrum. Meas.}
\bvolume{70},
\bfpage{1}--\blpage{12}
(\byear{2021})
\end{barticle}
\endbibitem

\bibitem[\protect\citeauthoryear{Kong et~al.}{2006}]{kong2006palmprint}
\begin{barticle}
\bauthor{\bsnm{Kong}, \binits{A.}},
\bauthor{\bsnm{Zhang}, \binits{D.}},
\bauthor{\bsnm{Kamel}, \binits{M.}}:
\batitle{Palmprint identification using feature-level fusion}.
\bjtitle{Pattern Recognit.}
\bvolume{39}(\bissue{3}),
\bfpage{478}--\blpage{487}
(\byear{2006})
\end{barticle}
\endbibitem

\bibitem[\protect\citeauthoryear{Fei et~al.}{2018}]{fei2018feature}
\begin{barticle}
\bauthor{\bsnm{Fei}, \binits{L.}},
\bauthor{\bsnm{Lu}, \binits{G.}},
\bauthor{\bsnm{Jia}, \binits{W.}},
\bauthor{\bsnm{Teng}, \binits{S.}},
\bauthor{\bsnm{Zhang}, \binits{D.}}:
\batitle{Feature extraction methods for palmprint recognition: A survey and
  evaluation}.
\bjtitle{IEEE Trans. Syst. Man Cybern. Syst.}
\bvolume{49}(\bissue{2}),
\bfpage{346}--\blpage{363}
(\byear{2018})
\end{barticle}
\endbibitem

\bibitem[\protect\citeauthoryear{Kong et~al.}{2003}]{kong2003palmprint}
\begin{barticle}
\bauthor{\bsnm{Kong}, \binits{W.K.}},
\bauthor{\bsnm{Zhang}, \binits{D.}},
\bauthor{\bsnm{Li}, \binits{W.}}:
\batitle{Palmprint feature extraction using 2-d gabor filters}.
\bjtitle{Pattern Recognit.}
\bvolume{36}(\bissue{10}),
\bfpage{2339}--\blpage{2347}
(\byear{2003})
\end{barticle}
\endbibitem

\bibitem[\protect\citeauthoryear{Jia et~al.}{2008}]{jia2008palmprint}
\begin{barticle}
\bauthor{\bsnm{Jia}, \binits{W.}},
\bauthor{\bsnm{Huang}, \binits{D.-S.}},
\bauthor{\bsnm{Zhang}, \binits{D.}}:
\batitle{Palmprint verification based on robust line orientation code}.
\bjtitle{Pattern Recognit.}
\bvolume{41}(\bissue{5}),
\bfpage{1504}--\blpage{1513}
(\byear{2008})
\end{barticle}
\endbibitem

\bibitem[\protect\citeauthoryear{Fei et~al.}{2016a}]{fei2016half}
\begin{barticle}
\bauthor{\bsnm{Fei}, \binits{L.}},
\bauthor{\bsnm{Xu}, \binits{Y.}},
\bauthor{\bsnm{Zhang}, \binits{D.}}:
\batitle{Half-orientation extraction of palmprint features}.
\bjtitle{Pattern Recognit. Lett.}
\bvolume{69},
\bfpage{35}--\blpage{41}
(\byear{2016})
\end{barticle}
\endbibitem

\bibitem[\protect\citeauthoryear{Fei et~al.}{2016b}]{fei2016double}
\begin{barticle}
\bauthor{\bsnm{Fei}, \binits{L.}},
\bauthor{\bsnm{Xu}, \binits{Y.}},
\bauthor{\bsnm{Tang}, \binits{W.}},
\bauthor{\bsnm{Zhang}, \binits{D.}}:
\batitle{Double-orientation code and nonlinear matching scheme for palmprint
  recognition}.
\bjtitle{Pattern Recognit.}
\bvolume{49},
\bfpage{89}--\blpage{101}
(\byear{2016})
\end{barticle}
\endbibitem

\bibitem[\protect\citeauthoryear{Jiang et~al.}{2023}]{jiang2023adversarial}
\begin{botherref}
\oauthor{\bsnm{Jiang}, \binits{F.}},
\oauthor{\bsnm{Li}, \binits{Q.}},
\oauthor{\bsnm{Liu}, \binits{P.}},
\oauthor{\bsnm{Zhou}, \binits{X.-D.}},
\oauthor{\bsnm{Sun}, \binits{Z.}}:
Adversarial learning domain-invariant conditional features for robust face
  anti-spoofing.
Int. J. Comput. Vis.,
1--24
(2023)
\end{botherref}
\endbibitem

\bibitem[\protect\citeauthoryear{Zhao et~al.}{2022}]{zhao2022structure}
\begin{botherref}
\oauthor{\bsnm{Zhao}, \binits{S.}},
\oauthor{\bsnm{Fei}, \binits{L.}},
\oauthor{\bsnm{Wen}, \binits{J.}},
\oauthor{\bsnm{Zhang}, \binits{B.}},
\oauthor{\bsnm{Zhao}, \binits{P.}},
\oauthor{\bsnm{Li}, \binits{S.}}:
Structure suture learning-based robust multiview palmprint recognition.
IEEE Trans. Neural Netw. Learn. Syst.
(2022)
\end{botherref}
\endbibitem

\bibitem[\protect\citeauthoryear{Matkowski
  et~al.}{2019}]{matkowski2019palmprint}
\begin{barticle}
\bauthor{\bsnm{Matkowski}, \binits{W.M.}},
\bauthor{\bsnm{Chai}, \binits{T.}},
\bauthor{\bsnm{Kong}, \binits{A.W.K.}}:
\batitle{Palmprint recognition in uncontrolled and uncooperative environment}.
\bjtitle{IEEE Trans. Inf. Forensics Secur.}
\bvolume{15},
\bfpage{1601}--\blpage{1615}
(\byear{2019})
\end{barticle}
\endbibitem

\bibitem[\protect\citeauthoryear{Chai et~al.}{2019}]{chai2019boosting}
\begin{barticle}
\bauthor{\bsnm{Chai}, \binits{T.}},
\bauthor{\bsnm{Prasad}, \binits{S.}},
\bauthor{\bsnm{Wang}, \binits{S.}}:
\batitle{Boosting palmprint identification with gender information using
  deepnet}.
\bjtitle{Future Gener. Comput. Syst.}
\bvolume{99},
\bfpage{41}--\blpage{53}
(\byear{2019})
\end{barticle}
\endbibitem

\bibitem[\protect\citeauthoryear{Genovese et~al.}{2019}]{genovese2019palmnet}
\begin{barticle}
\bauthor{\bsnm{Genovese}, \binits{A.}},
\bauthor{\bsnm{Piuri}, \binits{V.}},
\bauthor{\bsnm{Plataniotis}, \binits{K.N.}},
\bauthor{\bsnm{Scotti}, \binits{F.}}:
\batitle{Palmnet: Gabor-pca convolutional networks for touchless palmprint
  recognition}.
\bjtitle{IEEE Trans. Inf. Forensics Secur.}
\bvolume{14}(\bissue{12}),
\bfpage{3160}--\blpage{3174}
(\byear{2019})
\end{barticle}
\endbibitem

\bibitem[\protect\citeauthoryear{Zhong et~al.}{2018}]{zhong2018palm}
\begin{bchapter}
\bauthor{\bsnm{Zhong}, \binits{D.}},
\bauthor{\bsnm{Liu}, \binits{S.}},
\bauthor{\bsnm{Wang}, \binits{W.}},
\bauthor{\bsnm{Du}, \binits{X.}}:
\bctitle{Palm vein recognition with deep hashing network}.
In: \bbtitle{Proc. Chinese Conf. Pattern Recognit. Comput. Vis. (PRCV)},
pp. \bfpage{38}--\blpage{49}
(\byear{2018})
\end{bchapter}
\endbibitem

\bibitem[\protect\citeauthoryear{Zhu et~al.}{2016}]{zhu2016deep}
\begin{bchapter}
\bauthor{\bsnm{Zhu}, \binits{H.}},
\bauthor{\bsnm{Long}, \binits{M.}},
\bauthor{\bsnm{Wang}, \binits{J.}},
\bauthor{\bsnm{Cao}, \binits{Y.}}:
\bctitle{Deep hashing network for efficient similarity retrieval}.
In: \bbtitle{Proc. of the AAAI Conf. Artif. Intell. (AAAI)},
vol. \bseriesno{30}
(\byear{2016})
\end{bchapter}
\endbibitem

\bibitem[\protect\citeauthoryear{Wu et~al.}{2021}]{wu2021palmprint}
\begin{barticle}
\bauthor{\bsnm{Wu}, \binits{T.}},
\bauthor{\bsnm{Leng}, \binits{L.}},
\bauthor{\bsnm{Khan}, \binits{M.K.}},
\bauthor{\bsnm{Khan}, \binits{F.A.}}:
\batitle{Palmprint-palmvein fusion recognition based on deep hashing network}.
\bjtitle{IEEE Access}
\bvolume{9},
\bfpage{135816}--\blpage{135827}
(\byear{2021})
\end{barticle}
\endbibitem

\bibitem[\protect\citeauthoryear{Cho et~al.}{2019}]{cho2019extraction}
\begin{barticle}
\bauthor{\bsnm{Cho}, \binits{S.}},
\bauthor{\bsnm{Oh}, \binits{B.-S.}},
\bauthor{\bsnm{Toh}, \binits{K.-A.}},
\bauthor{\bsnm{Lin}, \binits{Z.}}:
\batitle{Extraction and cross-matching of palm-vein and palmprint from the rgb
  and the nir spectrums for identity verification}.
\bjtitle{IEEE Access}
\bvolume{8},
\bfpage{4005}--\blpage{4021}
(\byear{2019})
\end{barticle}
\endbibitem

\bibitem[\protect\citeauthoryear{Cho et~al.}{2021}]{cho2021palm}
\begin{barticle}
\bauthor{\bsnm{Cho}, \binits{S.}},
\bauthor{\bsnm{Oh}, \binits{B.-S.}},
\bauthor{\bsnm{Kim}, \binits{D.}},
\bauthor{\bsnm{Toh}, \binits{K.-A.}}:
\batitle{Palm-vein verification using images from the visible spectrum}.
\bjtitle{IEEE Access}
\bvolume{9},
\bfpage{86914}--\blpage{86927}
(\byear{2021})
\end{barticle}
\endbibitem

\bibitem[\protect\citeauthoryear{Su et~al.}{2023}]{su2023learning}
\begin{barticle}
\bauthor{\bsnm{Su}, \binits{L.}},
\bauthor{\bsnm{Fei}, \binits{L.}},
\bauthor{\bsnm{Zhao}, \binits{S.}},
\bauthor{\bsnm{Wen}, \binits{J.}},
\bauthor{\bsnm{Zhu}, \binits{J.}},
\bauthor{\bsnm{Teng}, \binits{S.}}:
\batitle{Learning modality-invariant binary descriptor for crossing palmprint
  to palm-vein recognition}.
\bjtitle{Pattern Recognit. Lett.}
\bvolume{172},
\bfpage{1}--\blpage{7}
(\byear{2023})
\end{barticle}
\endbibitem

\bibitem[\protect\citeauthoryear{Wicaksana
  et~al.}{2022}]{wicaksana2022customized}
\begin{barticle}
\bauthor{\bsnm{Wicaksana}, \binits{J.}},
\bauthor{\bsnm{Yan}, \binits{Z.}},
\bauthor{\bsnm{Yang}, \binits{X.}},
\bauthor{\bsnm{Liu}, \binits{Y.}},
\bauthor{\bsnm{Fan}, \binits{L.}},
\bauthor{\bsnm{Cheng}, \binits{K.-T.}}:
\batitle{Customized federated learning for multi-source decentralized medical
  image classification}.
\bjtitle{IEEE Journal of Biomedical and Health Informatics}
\bvolume{26}(\bissue{11}),
\bfpage{5596}--\blpage{5607}
(\byear{2022})
\end{barticle}
\endbibitem

\bibitem[\protect\citeauthoryear{McMahan
  et~al.}{2017}]{mcmahan2017communication}
\begin{bchapter}
\bauthor{\bsnm{McMahan}, \binits{B.}},
\bauthor{\bsnm{Moore}, \binits{E.}},
\bauthor{\bsnm{Ramage}, \binits{D.}}, \betal:
\bctitle{Communication-efficient learning of deep networks from decentralized
  data}.
In: \bbtitle{Artificial Intelligence and Statistics},
pp. \bfpage{1273}--\blpage{1282}
(\byear{2017}).
\bcomment{PMLR}
\end{bchapter}
\endbibitem

\bibitem[\protect\citeauthoryear{Li et~al.}{2020}]{li2020federated}
\begin{bchapter}
\bauthor{\bsnm{Li}, \binits{T.}},
\bauthor{\bsnm{Sahu}, \binits{A.K.}},
\bauthor{\bsnm{Zaheer}, \binits{M.}}, \betal:
\bctitle{Federated optimization in heterogeneous networks}.
In: \bbtitle{Proceedings of the Machine Learning and Systems},
vol. \bseriesno{2},
pp. \bfpage{429}--\blpage{450}
(\byear{2020})
\end{bchapter}
\endbibitem

\bibitem[\protect\citeauthoryear{Li et~al.}{2021}]{li2021fedbn}
\begin{botherref}
\oauthor{\bsnm{Li}, \binits{X.}},
\oauthor{\bsnm{Jiang}, \binits{M.}},
\oauthor{\bsnm{Zhang}, \binits{X.}}, et al.:
Fedbn: Federated learning on non-iid features via local batch normalization.
arXiv preprint arXiv:2102.07623
(2021)
\end{botherref}
\endbibitem

\bibitem[\protect\citeauthoryear{Arivazhagan
  et~al.}{2019}]{arivazhagan2019federated}
\begin{botherref}
\oauthor{\bsnm{Arivazhagan}, \binits{M.G.}},
\oauthor{\bsnm{Aggarwal}, \binits{V.}},
\oauthor{\bsnm{Singh}, \binits{A.K.}},
\oauthor{\bsnm{Choudhary}, \binits{S.}}:
Federated learning with personalization layers.
arXiv preprint arXiv:1912.00818
(2019)
\end{botherref}
\endbibitem

\bibitem[\protect\citeauthoryear{Wu et~al.}{2022}]{wu2022communication}
\begin{barticle}
\bauthor{\bsnm{Wu}, \binits{C.}},
\bauthor{\bsnm{Wu}, \binits{F.}},
\bauthor{\bsnm{Lyu}, \binits{L.}},
\bauthor{\bsnm{Huang}, \binits{Y.}},
\bauthor{\bsnm{Xie}, \binits{X.}}:
\batitle{Communication-efficient federated learning via knowledge
  distillation}.
\bjtitle{Nat. Commun.}
\bvolume{13}(\bissue{1}),
\bfpage{2032}
(\byear{2022})
\end{barticle}
\endbibitem

\bibitem[\protect\citeauthoryear{Liang et~al.}{2020}]{liang2020think}
\begin{botherref}
\oauthor{\bsnm{Liang}, \binits{P.P.}},
\oauthor{\bsnm{Liu}, \binits{T.}},
\oauthor{\bsnm{Ziyin}, \binits{L.}},
\oauthor{\bsnm{Allen}, \binits{N.B.}},
\oauthor{\bsnm{Auerbach}, \binits{R.P.}},
\oauthor{\bsnm{Brent}, \binits{D.}},
\oauthor{\bsnm{Salakhutdinov}, \binits{R.}},
\oauthor{\bsnm{Morency}, \binits{L.-P.}}:
Think locally, act globally: Federated learning with local and global
  representations.
arXiv preprint arXiv:2001.01523
(2020)
\end{botherref}
\endbibitem

\bibitem[\protect\citeauthoryear{Yang et~al.}{2022}]{yang2022hypernetwork}
\begin{botherref}
\oauthor{\bsnm{Yang}, \binits{Z.}},
\oauthor{\bsnm{Xia}, \binits{W.}},
\oauthor{\bsnm{Lu}, \binits{Z.}},
\oauthor{\bsnm{Chen}, \binits{Y.}},
\oauthor{\bsnm{Li}, \binits{X.}},
\oauthor{\bsnm{Zhang}, \binits{Y.}}:
Hypernetwork-based personalized federated learning for multi-institutional ct
  imaging.
arXiv preprint arXiv:2206.03709
(2022)
\end{botherref}
\endbibitem

\bibitem[\protect\citeauthoryear{Shao and Zhong}{2020}]{shao2020towards}
\begin{barticle}
\bauthor{\bsnm{Shao}, \binits{H.}},
\bauthor{\bsnm{Zhong}, \binits{D.}}:
\batitle{Towards privacy palmprint recognition via federated hash learning}.
\bjtitle{Electron. Lett.}
\bvolume{56}(\bissue{25}),
\bfpage{1418}--\blpage{1420}
(\byear{2020})
\end{barticle}
\endbibitem

\bibitem[\protect\citeauthoryear{DeJonge et~al.}{2016}]{dejonge2016assessing}
\begin{barticle}
\bauthor{\bsnm{DeJonge}, \binits{K.C.}},
\bauthor{\bsnm{Mefford}, \binits{B.S.}},
\bauthor{\bsnm{Ch{\'a}vez}, \binits{J.L.}}:
\batitle{Assessing corn water stress using spectral reflectance}.
\bjtitle{International Journal of Remote Sensing}
\bvolume{37}(\bissue{10}),
\bfpage{2294}--\blpage{2312}
(\byear{2016})
\end{barticle}
\endbibitem

\bibitem[\protect\citeauthoryear{Khosla et~al.}{2020}]{khosla2020supervised}
\begin{bchapter}
\bauthor{\bsnm{Khosla}, \binits{P.}},
\bauthor{\bsnm{Teterwak}, \binits{P.}},
\bauthor{\bsnm{Wang}, \binits{C.}},
\bauthor{\bsnm{Sarna}, \binits{A.}},
\bauthor{\bsnm{Tian}, \binits{Y.}},
\bauthor{\bsnm{Isola}, \binits{P.}},
\bauthor{\bsnm{Maschinot}, \binits{A.}},
\bauthor{\bsnm{Liu}, \binits{C.}},
\bauthor{\bsnm{Krishnan}, \binits{D.}}:
\bctitle{Supervised contrastive learning}.
In: \bbtitle{Proc. Adv. Neural Inf. Process. Syst. (NIPS)},
vol. \bseriesno{33},
pp. \bfpage{18661}--\blpage{18673}
(\byear{2020})
\end{bchapter}
\endbibitem

\bibitem[\protect\citeauthoryear{Kingma and Ba}{2014}]{kingma2014adam}
\begin{botherref}
\oauthor{\bsnm{Kingma}, \binits{D.P.}},
\oauthor{\bsnm{Ba}, \binits{J.}}:
Adam: A method for stochastic optimization.
arXiv preprint arXiv:1412.6980
(2014)
\end{botherref}
\endbibitem

\bibitem[\protect\citeauthoryear{Jain et~al.}{2004}]{JainBiom}
\begin{barticle}
\bauthor{\bsnm{Jain}, \binits{A.K.}},
\bauthor{\bsnm{Ross}, \binits{A.}},
\bauthor{\bsnm{Prabhakar}, \binits{S.}}:
\batitle{An introduction to biometric recognition}.
\bjtitle{IEEE Trans. Circuits Syst. for Video Technol.}
\bvolume{14}(\bissue{1}),
\bfpage{4}--\blpage{20}
(\byear{2004})
\end{barticle}
\endbibitem

\bibitem[\protect\citeauthoryear{Hao et~al.}{2008}]{hao2008multispectral}
\begin{bchapter}
\bauthor{\bsnm{Hao}, \binits{Y.}},
\bauthor{\bsnm{Sun}, \binits{Z.}},
\bauthor{\bsnm{Tan}, \binits{T.}},
\bauthor{\bsnm{Ren}, \binits{C.}}:
\bctitle{Multispectral palm image fusion for accurate contact-free palmprint
  recognition}.
In: \bbtitle{2008 15th IEEE International Conference on Image Processing},
pp. \bfpage{281}--\blpage{284}
(\byear{2008}).
\bcomment{IEEE}
\end{bchapter}
\endbibitem

\end{thebibliography}

\end{document}